\documentclass[pmlr,twocolumn,10pt]{jmlr} % W&CP article

\usepackage{xurl}
\usepackage{booktabs}
\usepackage{siunitx}
\usepackage{multirow}

\usepackage[switch]{lineno}

\theorembodyfont{\upshape}
\theoremheaderfont{\scshape}
\theorempostheader{:}
\theoremsep{\newline}

\jmlrvolume{333}
\jmlryear{2026}
\jmlrsubmitted{LEAVE UNSET}
\jmlrpublished{LEAVE UNSET}
\jmlrworkshop{Conference on Health, Inference, and Learning (CHIL) 2026} % W&CP title

\title[Benchmark Study of LLM Agents for Multimodal Clinical Prediction Tasks]{AgentRx: A Benchmark Study of LLM Agents for Multimodal Clinical Prediction Tasks}

\author{%
\Name{Baraa Al Jorf} \Email{baraa.al.jorf@nyu.edu}\\
\addr New York University Abu Dhabi, UAE\\
\Name{Farah E. Shamout} \Email{fs999@nyu.edu}\\
\addr New York University Abu Dhabi, UAE
}

\begin{document}

\maketitle

\begin{abstract}
Building effective clinical decision support systems requires the synthesis of complex heterogeneous multimodal data. Such modalities include temporal electronic health records data, medical images, radiology reports, and clinical notes. Large language model (LLM)–based agents have shown impressive performance in various healthcare tasks, especially those involving textual modalities. Considering the fragmentation of healthcare data across hospital systems, collaborative agent frameworks present a promising direction to mitigate data sharing challenges. However, the effectiveness of LLM agents for multimodal clinical risk prediction remains largely unexamined. In this work, we conduct a systematic evaluation of LLM-based agents for clinical prediction tasks using large-scale real-world data. We assess performance in unimodal and multimodal settings and quantify performance gaps between single agent and multi-agent systems. Our findings highlight that single agent frameworks outperform naive multi-agent systems, are better at handling multimodal data, and are better calibrated. This underscores a critical need for improving multi-agent collaboration to better handle heterogeneous inputs. By open-sourcing our code and evaluation framework, this work offers a new benchmark to support future developments relating to agentic systems in healthcare. 
%in-hospital mortality and length-of-stay prediction. \texttt{AgentRx} assesses performance across unimodal and multimodal settings. We quantify the performance gap between single agent systems, multi-agent systems, and specialized deep learning models.
%We highlight the current performance and robustness of multi-agent systems, and underscore the critical need for improved calibration in agentic healthcare workflows.
\end{abstract}

\paragraph*{Data and Code Availability}
This study utilizes publicly available data from MIMIC-IV \citep{johnson_mimic-iv_2023}, MIMIC-CXR \citep{johnson_mimic-cxr_2019}, and MIMIC-IV-Note \citep{johnson_mimic-iv-note_2023}. Our code and models are open source at: \url{https://github.com/nyuad-cai/AgentRX}.

\paragraph*{Institutional Review Board (IRB)}
This work did not involve human subjects, so IRB approval was not required.

\section{Introduction}
\label{sec:intro}
The integration of Artificial Intelligence (AI) into clinical decision support systems promises more optimized clinical workflows and better patient outcomes, particularly in resource-constrained settings such as the Intensive Care Unit (ICU). There are many data modalities routinely collected from patients in the ICU, such as Patient Summaries (PS), Electronic Health Records (EHR) data, Chest X-ray (CXR) images, Radiology Reports (RR), and Discharge Notes (DN). Recent work has highlighted the potential of multimodal deep neural networks for fusing such modalities to more accurately predict patient risk outcomes, compared to relying on a single modality \citep{khader_medical_2023, lee_learning_2023}.

State-Of-The-Art (SOTA) prediction models rely on optimized deep learning-based architectures. For example, MeTra uses a transformer to apply an attention-based fusion mechanism for processing EHR and CXR data \citep{khader_medical_2023}. Similarly, MedFuse uses a simpler Long Short-Term Memory (LSTM) network to fuse latent embeddings of the same modalities \citep{hayat_medfuse_2022}. MedPatch processes EHR, CXR, RR, and DN data, and uses a modular multistage fusion pipeline \citep{al_jorf_medpatch_2025}. While the models achieve high performance metrics in clinical prediction tasks, they have some limitations  that impede real-world adoption. First, they are black-box in nature, which undermines clinical trust and interpretability \citep{catalina_knowledge_2023,von_eschenbach_transparency_2021,shuaib_transforming_2024}. Second, they typically have fixed data requirements, which restrict the transferability of these approaches across other modalities and clinical settings \citep{al_jorf_data-centric_2026}. %Consequently, they are classified as static architectures with limited deployment capabilities.

Recently, LLMs have demonstrated powerful abilities in emulating human reasoning and text generation \citep{singhal_toward_2025} and have proven to enhance clinical workflows \citep{tai-seale_ai-generated_2024}, making them attractive candidates for clinical decision support tools. Recent work specifically highlights their effective ability in generating narrative justifications \citep{lee_prompt_2025}. Unlike traditional black box models that rely on abstract feature importance, LLMs can function as semantic translators, converting complex risk predictions into intuitive, natural language explanations that align with clinical reasoning. However, this advantage comes with unreliable performance gains. While some studies highlight the effectiveness of LLMs at handling unimodal EHR or PS data for clinical prediction tasks \citep{acharya_clinical_2024, kara_clinical_2025, jin_agentmd_2025}, others find that LLMs underperform compared to traditional supervised methods \citep{tan_are_2024, zhu_medagentboard_2025}. Notably, these studies have been restricted to unimodal settings. To date, the potential of LLMs to handle multimodal data integration for robust clinical risk prediction remains unexplored.

% However, this advantage comes with a critical performance caveat:  LLMs have been shown to significantly underperform compared to traditional specialized machine learning models in risk prediction and forecasting tasks \citep{kara_clinical_2025,rezk_llms_2024,tan_are_2024}. This disparity defines a critical gap in clinical AI, where the potential advantages of  LLMs are not fully realized due to suboptimal system performance.

% Paragraph 4: Introducing the study's primary research question and motivation. 
To address this gap, this work investigates the fundamental question: How effective are agentic systems at multimodal clinical prediction tasks? Our motivation stems from the fragmented nature of healthcare data, where clinical information is often dispersed across multiple isolated databases that ideally would be managed by modality-specific agents to avoid expensive data transfers. However, it remains unclear whether such decentralized, agentic approaches perform well across a variety of data availability settings. Hence, we introduce \texttt{AgentRx}, a comprehensive evaluation framework to analyze agent performance across three progressive settings. First, we establish performance baselines using only a single modality, specifically clinical notes encompassed within patient summaries. Second, we assess the capacity of a single agent to synthesize heterogeneous multimodal data within one context window, utilizing modality dropping ablations to measure robustness. Finally, we investigate whether multi-agent reasoning across specialized agents can improve performance compared to single agent approaches.

% We therefore seek to rigorously benchmark these architectures to understand how agentic reasoning scales with multiple modalities.

% % Paragraph 5: Explicitly framing the experiments around the three settings: Unimodal, Multimodal Single, and Multimodal Multi-Agent.
% Paragraph 6: Summarizing the key contributions.
In summary, we make the following contributions: 
\begin{itemize} 
\item We provide a systemic benchmark (\texttt{AgentRx}) for the evaluation of LLM-based clinical prediction tasks using four data modalities: EHR, RR, CXR, and PS.  %derived from MIMIC-IV, MIMIC-CXR, and MIMIC-Note.
\item We analyze the impact of data heterogeneity on model performance, in terms of discriminative ability and calibration, by varying the number and type of modalities available to the agents. We also conduct ablations to assess the effect of progressively adding more data modalities in simple single agent and multi-agent setups, to quantify the performance gap between the two.
% .. TODO 
\item We publicly release our code and evaluation framework to enhance the usability of \texttt{AgentRx} by the research community and support the advancement of agentic AI in healthcare.
\end{itemize}

\begin{table*}[ht]
    \caption{{\textbf{Summary of existing medical multimodal benchmarks for risk prediction.}} We summarize the characteristics of existing benchmarks, including both traditional deep learning benchmarks and agentic frameworks. The ``Agentic" column indicates whether the benchmark explicitly evaluates agent-based reasoning workflows.}
    \centering
    \setlength{\tabcolsep}{4pt} % Slightly increased spacing since we removed columns
    \resizebox{0.9\linewidth}{!}{
    \begin{tabular}{l c c c } \toprule

         \textbf{Benchmark} & \textbf{Modalities} & \textbf{Scope} & \textbf{Agentic} \\
        
        \midrule
         % Non-Agentic Baselines
         DF-Mdl~\citep{chen_multi-modal_2024} & PS, Medical Images & Clinical Prediction  & $\times$\\   

         MC-BEC~\citep{chen_multimodal_2023} & PS, EHR & Clinical Prediction  & $\times$\\   
         
         MedPatch~\citep{al_jorf_medpatch_2025} & EHR, CXR, RR, DN & Clinical Prediction  & $\times$\\

         MedMod~\citep{elsharief_medmod_2025} & EHR, CXR & Clinical Prediction & $\times$\\
          
        \midrule
        % Agentic Benchmarks
            
         EHRXQA~\citep{bae_ehrxqa_2023} & EHR, CXR & Medical Question Answering & $\checkmark$\\ 

         MedAgents~\citep{tang_medagents_2024} & Medical Questions & Medical Question Answering  & \checkmark \\

         MDAgents~\citep{kim_mdagents_2024} & Medical Questions, Images, Videos & Medical Question Answering & \checkmark \\
          \midrule
           \textbf{AgentRx} (Ours) & EHR, CXR, RR, PS & \textbf{Clinical Prediction} & \checkmark\\
         
        \bottomrule
    \end{tabular}}
    \label{tab:benchmarks_comparison}
\end{table*}

\section{Related Work}
\subsection{LLMs in Healthcare}
The application of Natural Language Processing (NLP) in healthcare has rapidly evolved from task-specific discriminative models to general-purpose generative reasoning models. Early approaches relied on encoder-only architectures like BioBERT \citep{lee_biobert_2020} and ClinicalBERT \citep{huang_clinicalbert_2019} for medical tasks like biomedical text mining and hospital readmission prediction. The emergence of LLMs fundamentally shifted this focus towards generative capabilities. Models such as GPT-4 \citep{bicknell_chatgpt-4_2024} and Med-PaLM \citep{singhal_large_2023} achieved expert-level performance on USMLE-style reasoning and medical question answering. However, while LLMs excel at reasoning tasks like summarizing clinical notes \citep{afshar_pragmatic_2025, lukac_ambient_2025} and structuring interpretable clinical insights \citep{lee_prompt_2025}, their zero-shot ability to forecast temporal clinical outcomes often lags behind traditional supervised baselines \citep{tan_are_2024,zhu_medagentboard_2025}. This performance gap suggests that scaling model parameters alone is insufficient for risk stratification, necessitating the development of more structured reasoning frameworks and the integration of multimodal clinical data.

Consequently, Vision-Language Models (VLMs) extend LLM capabilities by aligning visual encoders with language decoders. In the general domain, architectures like Ovis \citep{lu2025ovis25technicalreport} and Qwen2.5-VL \citep{bai_qwen25-vl_2025} have demonstrated that visual signals can be effectively tokenized and processed alongside text. Within healthcare, this has led to specialized models such as LLaVA-Med \citep{li_llava-med_2023}, MedGemma \citep{sellergren_medgemma_2025}, and HuatuoGPT-Vision \citep{chen_huatuogpt-vision_2024}, which are fine-tuned on biomedical image-text pairs to perform tasks like visual question answering and instruction-following. Similar to their text-only counterparts, medical VLMs are predominantly evaluated on and designed for reasoning tasks rather than predictive ones \citep{kalpelbe_vision_2025}.

\subsection{Health Reasoning Agents using LLMs}
Since full parameter fine-tuning is often computationally expensive, recent research has pivoted towards optimizing inference-time reasoning to enhance clinical performance. Standard generic techniques have been developed to enhance single agent LLM reasoning. This includes Chain-of-Thought (CoT) prompting \citep{wei_chain--thought_2022} where the model is allowed to decompose a problem into manageable intermediate steps, Self-Consistency (SC) \citep{wang_self-consistency_2022} where multiple reasoning paths are used to mitigate errors, Retrieval-Augmented Generation (RAG) where LLMs are provided factual grounding from external knowledge bases \citep{lewis_retrieval-augmented_2020}, and Self-Refinement, which employs iterative critique to correct generated outputs based on feedback \citep{madaan_self-refine_2023}. These general-purpose strategies have been extensively benchmarked within the medical domain to validate their utility across tasks ranging from medical exams to clinical prediction \citep{tan_are_2024}.

Building on these foundations, domain-specific frameworks have been designed to further optimize medical reasoning. A prominent example is Medprompt \citep{nori_can_2023}, which synergizes RAG based few-shot selection, CoT reasoning chains, and ensembling to achieve SOTA performance on the MedQA benchmark. Expanding on this, AgentMD \citep{jin_agentmd_2025} introduces a tool-learning framework that empowers agents to autonomously curate and apply executable clinical calculators, demonstrating significantly improved accuracy in risk prediction compared to standard LLMs. While effective, these strategies primarily focus on optimizing the output of a single model instance.

\subsection{Multi-Agent Systems}
Numerous studies have demonstrated that employing multi-agent frameworks can significantly enhance LLM performance on complex reasoning tasks compared to single-agent baselines \citep{hong_metagpt_2023, yang_agentnet_2025}. By simulating human-like collaboration, these systems mitigate individual hallucinations and improve logical consistency. For instance, debate-style frameworks allow agents to critique each other's outputs until a consensus is reached, effectively filtering out erroneous reasoning steps \citep{liang_encouraging_2024, du_improving_2024}. In the medical domain, architectures such as MedAgents \citep{tang_medagents_2024} and MDAgents \citep{kim_mdagents_2024} have successfully applied role-playing strategies where agents adopt specific specialist personas to achieve SOTA results on medical question-answering benchmarks. However, preliminary literature suggests that this reasoning advantage may not translate directly to all domains  \citep{gao_single-agent_2025, kim_towards_2025, cemri_why_2025}. For clinical risk prediction, recent benchmarks indicate that while agents excel at generating explanations, they frequently underperform compared to specialized supervised models in temporal forecasting tasks \citep{zhu_medagentboard_2025, tan_are_2024}. Crucially, these negative findings have largely been derived from limited unimodal evaluations, primarily focusing on structured EHR data or text-only inputs. It remains unclear whether the collaborative benefits of multi-agent systems could be better realized in a multimodal setting, where distinct specialized agents could independently process heterogeneous data before engaging in collective decision-making.

\subsection{Motivation} 
We provide an overview of relevant benchmarks in Table \ref{tab:benchmarks_comparison}. Existing benchmarks largely focus on either traditional deep learning models or modern generative reasoning. Most deep learning based benchmarks focus on clinical prediction tasks, such as in-hospital mortality, length of stay, or patient phenotyping \citep{bae_ehrxqa_2023, elsharief_medmod_2025}, while agentic benchmarks focused mostly on reasoning and retrieval capabilities for medical question and answering. For example, MedAgents and MDAgents have established rigorous standards for multimodal medical LLMs, but predominantly focused on Medical Question Answering \citep{tang_medagents_2024, kim_mdagents_2024}. While current LLM medical benchmarks evaluate an agent's ability to retrieve knowledge or answer complex queries, they do not assess the agent's capacity for temporal forecasting or risk stratification in a clinical multimodal setting. The limited agentic prediction benchmarks that do exist are strictly unimodal \citep{tan_are_2024,zhu_medagentboard_2025}. This distinction highlights a critical need for benchmarks that specifically evaluate VLMs on predictive clinical endpoints, which is the main focus of our study.

\section{Methodology}
\label{sec:methods}

\subsection{Preliminaries}
To formalize the \texttt{AgentRx} benchmark, we introduce relevant notation. For a given patient encounter $p$, we assume the presence of multimodal data $\mathcal{X}_p = \{ \mathbf{x}_{ps}, \mathbf{x}_{ehr}, \mathbf{x}_{cxr}, \mathbf{x}_{rr} \}$. 
Here, $\mathbf{x}_{ps}$ represents a textual patient summary which includes the patient age, sex, and clinical history, $\mathbf{x}_{ehr} \in \mathbb{R}^{d \times t}$ represents the multivariate time-series data from the patient's EHR with $d$ features over $t$ time steps. $\mathbf{x}_{cxr} \in \mathbb{R}^{h \times w}$ represents a CXR image with height and width $h, w = 224$ and $\mathbf{x}_{rr}$ represents the textual radiology reports associated with all medical images collected from the patient. The goal of the system is to predict a groundtruth label denoted by $y$. 

% \begin{figure}[t]
%     \centering
% \includegraphics[width=1.0\linewidth]{figures/Data.png} \vspace{-5mm}
%     \caption{Dataset distribution across the training, validation, and testing splits for both Mortality and Length-of-Stay tasks.}\vspace{-5mm}
%     \label{Figure2}
% \end{figure}

\begin{table}[t!]
    \caption{{\textbf{Dataset statistics per modality and task.}} We describe the sample counts for the training and test splits across the mortality and length of stay prediction tasks. The counts are stratified by modality.}
    \centering
    \setlength{\tabcolsep}{8pt} % Adjusted spacing for better readability
    \resizebox{0.9\linewidth}{!}{
    \begin{tabular}{l c c c c } \toprule
         & \multicolumn{2}{c}{\textbf{Mortality}} & \multicolumn{2}{c}{\textbf{Length of Stay}} \\
         \cmidrule(lr){2-3} \cmidrule(lr){4-5}
         \textbf{Modality} & \textbf{Training} & \textbf{Test} & \textbf{Training} & \textbf{Test} \\
        \midrule
         PS & 17,773 & 4,925 & 17,476 & 4,845 \\
         EHR & 17,773 & 4,925 & 17,476 & 4,845 \\
         RR  & 16,128 & 4,454 & 15,865 & 4,380 \\
         CXR  & 4,259 & 1,174 & 4,171 & 1,153 \\
        \bottomrule
    \end{tabular}}
    \label{tab:dataset_stats}
\end{table}

\subsection{Real-world Multimodal Dataset}
\subsubsection{Data Curation}
We extracted the EHR data from MIMIC-IV \citep{johnson_mimic-iv_2023}, the CXR images from MIMIC-CXR \citep{johnson_mimic-cxr_2019}, and the RR and PS from MIMIC-IV-Notes \citep{johnson_mimic-iv-note_2023}. The MIMIC-IV dataset includes de-identified data from over $315,460$ patients with ICU stays at the Beth Israel Deaconess Medical Center between 2008 and 2019. The MIMIC-CXR dataset consists of over $377,000$ chest radiographs. MIMIC-IV-Note supplements these two datasets with unstructured textual data, including both 331,794 DNs and 2,321,355 RRs. We constructed the multimodal dataset by aligning the subject, stay, and admission identifiers across all three datasets.  We mandate the existence of PS for all samples, while other modalities are paired if available.

\subsubsection{Prediction Tasks}
We introduce two clinical prediction tasks and perform evaluations across two modality settings, with a
unified evaluation scheme. We report the Area Under the Receiver Operating Characteristic Curve (AUROC), Area Under the Precision-Recall Curve (AUPRC), and Expected Calibration Error (ECE):
\begin{itemize}
    \item In-hospital mortality prediction is a binary classification task that involves predicting risk of in-hospital mortality based on the first 48 hours in the ICU.
    \item Long Length of Stay prediction is a binary classification task that involves predicting whether a patient's stay is going to extend beyond 7 days, by the end of the first 48 hours in the ICU.
\end{itemize}
After multimodal pairing, the dataset was split into training (70\%), validation (10\%), and testing (20\%) sets. The exact dataset distribution per task is reported in Table \ref{tab:dataset_stats}.

%We focus on two main clinical prediction tasks: in-hospital morality prediction and length of stay prediction using the first 48 hours of an ICU stay. Hence, we applied an inclusion and exclusion criteria where only patients with an ICU stay greater than 48 hours were considered. Additionally, we only consider the modalities collected before the 48 hour mark.  

\begin{figure*}[ht]
    \centering
    \includegraphics[width=\textwidth]{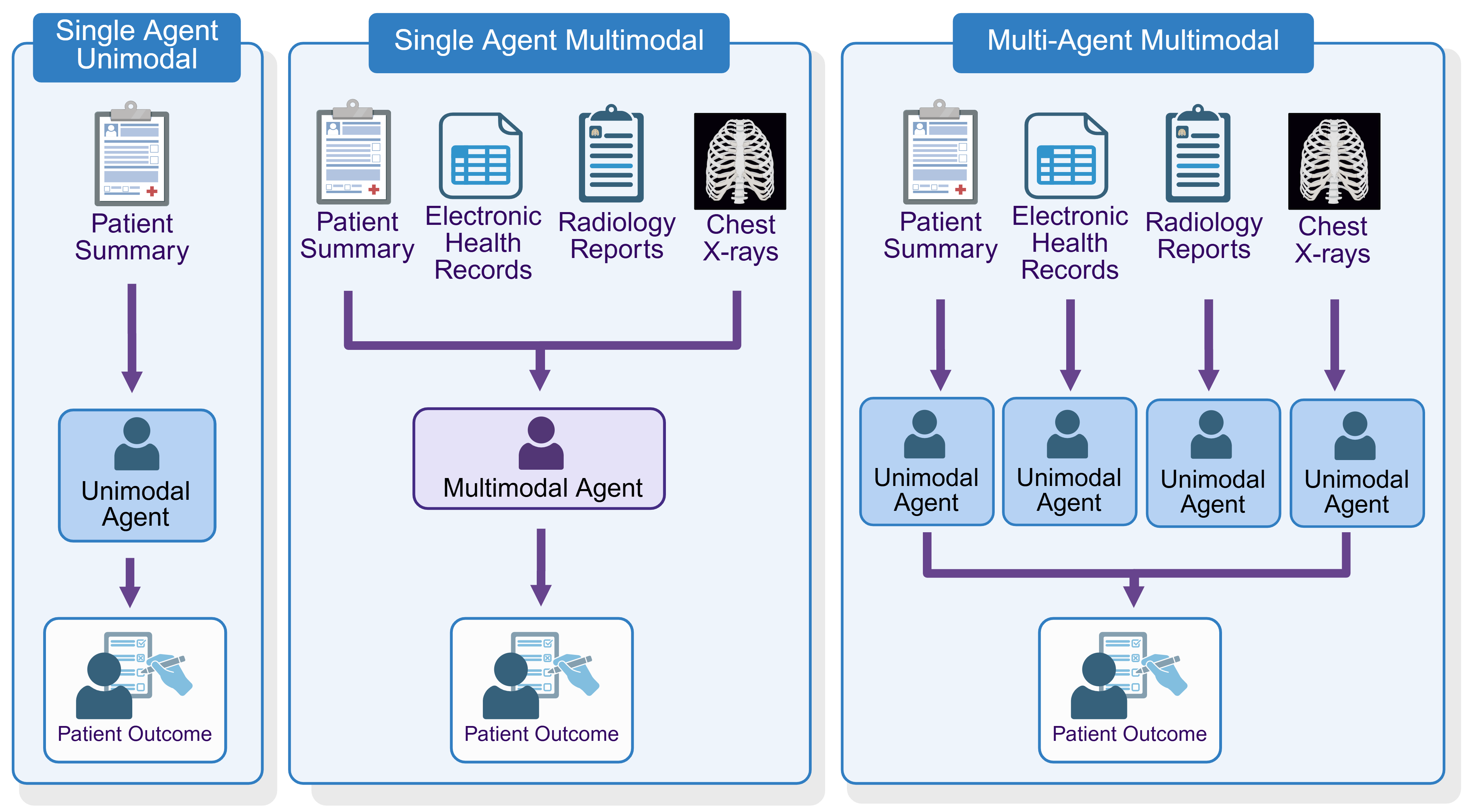}
    \caption{Overview of the agentic evaluation frameworks considered within the \texttt{AgentRx} benchmark, spanning Single-Agent (Unimodal/Multimodal) and Multi-Agent settings. }\vspace{-5mm}
    \label{Figure1}
\end{figure*}

\subsubsection{Multimodal Data Processing and Pairing}
Since patient summaries are considered to be the primary base modality, we ensure that every patient in our cohort has one. We generated patient summaries by processing the raw DNs from MIMIC-IV-Note. To prevent data leakage, we extracted only the information available prior to ICU admission. This includes the patient's medical, surgical, and family histories as well as the demographic details such as age and sex, ensuring the agent operates with the same context available to a clinician at the time of admission. 

% \subsubsection{EHR Data Preprocessing}
For EHR, we used a set of 17 clinical variables consistent with previous work \citep{hayat_medfuse_2022, al_jorf_medpatch_2025}. These include 5 categorical variables (capillary refill rate, Glasgow coma scale eye opening, motor response, verbal response, and total) and 12 continuous variables (diastolic blood pressure, fraction of inspired oxygen, glucose, heart rate, height, mean blood pressure, oxygen saturation, respiratory rate, systolic blood pressure, temperature, weight, and pH). To format the EHR data for the LLM agents, we serialized it into a structured text format \texttt{[$T_{0}$+$\Delta T$] Variable=Value} where $T_{0}$ is the time of admission, recording only observed measurements at each timestamp to minimize token usage and handle irregular sampling rates efficiently. The observation window was strictly limited to the first 48 hours of ICU admission. To strictly adhere to context window limits, high-frequency streams exceeding 500 time-steps were truncated to preserve the initial admission state (first 100 steps) and the most recent clinical trajectory (last 400 steps).

% \subsubsection{Chest X-ray Images}
For the same cohort, we included CXR images that were collected within the first 48 hours of ICU admission. We restricted the selection to Anterior-Posterior (AP) views, as these are standard for portable ICU bedside imaging. For patients with multiple scans within the window, we selected the latest valid scan to capture the most recent clinical state prior to the prediction horizon. 

% Anything interesting here relating to agents?

% \subsubsection{Radiology Reports}
We extracted reports corresponding to patients available in the dataset. MIMIC-IV-Note includes reports for various imaging modalities (CT, MRI, Ultrasound). We concatenated all relevant reports for each patient that were collected within 48 hours from admission into a single report aggregate that is passed as the full RR modality. 
\subsection{Agentic Frameworks}
We formalize the inference logic for our agentic evaluation framework. We define an agent $\mathcal{A}(\cdot)$ as a functional unit that processes clinical context to produce a prediction or a probability. We define $\mathcal{P}_{task}$ as the system prompt directing the agent's reasoning for unimodal and multimodal tasks, and explore three main settings depicted in Figure~\ref{Figure1}:
\begin{itemize}
    \item Setting 1: Single Agent Unimodal \\
In this setting, the agent relies solely on the Patient Summary ($\mathbf{x}_{ps}$) to form a baseline prediction as demonstrated in Appendix \ref{alg:single_unimodal}.

    \item Setting 2: Single Agent Multimodal \\
    Here, a single generalist agent processes all available modalities $\mathcal{X}_{available}$ simultaneously within a single context window as demonstrated in Appendix \ref{alg:single_multimodal}.

    \item Setting 3: Multi-Agent Multimodal \\
This setting employs specialized agents $\mathcal{A}_m$ for each modality $m$. Each agent independently generates a probability $p_m$, which is then averaged. This is shown in Appendix \ref{alg:multi_agent}

\end{itemize}

\subsection{Baselines}
To assess the efficacy of different agentic architectures, we compare performance across the following single-agent and multi-agent baselines. %The architectural setups are depicted in Figure \ref{Figure1}.

\subsubsection{Supervised Baselines}
We compare our agentic frameworks against two specialized deep learning architectures:
\begin{enumerate}
    \item BioBERT (Unimodal): For the text-only setting, we utilize BioBERT \citep{lee_biobert_2020}, a language model pre-trained on biomedical corpora (PubMed). We freeze the backbone and fine-tune a linear classification head on the token embeddings of the PS to establish a strong supervised baseline.
    \item MedPatch (Multimodal): For the multimodal setting, we employ MedPatch \citep{al_jorf_medpatch_2025}, a SOTA fusion architecture that utilizes a confidence-guided patching mechanism to effectively integrate heterogeneous modalities.
\end{enumerate}
\subsubsection{Single-Agent Baselines - Unimodal and Multimodal}
\begin{enumerate}
    \item Zero-shot: Vanilla baseline where we feed the model all the setting's available modalities and ask it for a prediction.
    \item Few-shot: A baseline where we feed the model one positive example and one negative example (data + labels from the training set) and then ask it to predict the outcome for a test sample \citep{brown_language_2020}.
    \item Chain-of-Thought (CoT): A baseline where we allow the model to first generate a reasoning step, then use that reasoning to produce a prediction \citep{wei_chain--thought_2022}.
    \item Self-Consistency + Chain-of-Thought (CoT-SC): A baseline where the model runs 3 parallel reasoning pathways and then makes a decision using all three paths via voting (probability averaging) \citep{wang_self-consistency_2022}.
    \item Self-Refinement: A baseline where the model generates a prediction with reasoning, then self-evaluates its reasoning before making a final prediction based on the evaluation feedback \citep{madaan_self-refine_2023}.
\end{enumerate}

\subsubsection{Multi-Agent Baselines}
\begin{enumerate}
    \item Majority Vote: A baseline where unimodal agents independently analyze their specific data and vote on the outcome \citep{kaesberg_voting_2025}.
    \item Debate: A baseline where unimodal agents debate with each other until they reach a consensus \citep{du_improving_2024}.
    \item Meta-Prompting: A baseline where a meta-agent evaluates the data and either makes a prediction or instantiates other expert agents to help refine its task \citep{hou_metaprompting_2022}.
    \item Traj-CoA + Multimodal Judge: A baseline that uses multiple worker agents to construct an EHR memory. Each worker receives a chunk of EHR data. The EHR memory and the other modalities are then passed to a multimodal judge agent that makes a final prediction. This baseline combines a unimodal and a multimodal agent setup \citep{zeng_traj-coa_nodate}.
    \item MDAgents: A baseline where a diverse ensemble of agents is initialized depending on patient case severity \citep{kim_mdagents_2024}.
    \item MedAgents: A baseline where agents follow predefined roles and collaborate to create a patient state summary to enable outcome prediction  \citep{tang_medagents_2024}.
\end{enumerate}

\begin{table*}[ht]
\caption{\textbf{Single Agent Unimodal Results.} Comparison of agentic backbones against a supervised baseline (BioBERT) using only the PS. Metrics are reported with 95\% Confidence Intervals. The best overall performance in each column is bolded.}
\label{tab:unimodal}
\centering
\resizebox{\linewidth}{!}{%
\begin{tabular}{llcccccc}
\toprule
\multirow{2}{*}{\textbf{Backbone}} & \multirow{2}{*}{\textbf{Method}} & \multicolumn{3}{c}{\textbf{In-Hospital Mortality}} & \multicolumn{3}{c}{\textbf{Length of Stay ($>$7 Days)}} \\
\cmidrule(lr){3-5} \cmidrule(lr){6-8}
 & & \textbf{AUROC} & \textbf{AUPRC} & \textbf{ECE} & \textbf{AUROC} & \textbf{AUPRC} & \textbf{ECE} \\
\midrule
Supervised & BioBERT & 0.680 (0.657 - 0.702) & 0.228 (0.203 - 0.263) & \textbf{0.006} & \textbf{0.641 (0.621 - 0.661)} & \textbf{0.307 (0.282 - 0.337)} & \textbf{0.024} \\
\midrule
\multirow{5}{*}{\textbf{Qwen}}  & Zero-shot & 0.667 (0.645 - 0.690) & 0.234 (0.208 - 0.267) & 0.025 & 0.603 (0.582 - 0.624) & 0.261 (0.241 - 0.286) & 0.105 \\
 & Few-shot & 0.697 (0.678 - 0.718) & 0.236 (0.213 - 0.270) & 0.042 & 0.602 (0.580 - 0.621) & 0.257 (0.237 - 0.280) & 0.122 \\
 & CoT & 0.650 (0.629 - 0.675) & 0.214 (0.191 - 0.244) & 0.060 & 0.592 (0.573 - 0.613) & 0.260 (0.240 - 0.286) & 0.480 \\
 & CoT-SC & 0.679 (0.657 - 0.702) & 0.236 (0.209 - 0.269) & 0.029 & 0.593 (0.571 - 0.613) & 0.259 (0.239 - 0.284) & 0.454 \\
 & Self-Refine & 0.666 (0.644 - 0.689) & 0.212 (0.187 - 0.240) & 0.080 & 0.580 (0.560 - 0.601) & 0.236 (0.219 - 0.257) & 0.198 \\
\midrule
\multirow{5}{*}{\textbf{Intern}}  & Zero-shot & 0.672 (0.650 - 0.695) & 0.216 (0.193 - 0.245) & 0.090 & 0.620 (0.599 - 0.641) & 0.278 (0.256 - 0.304) & 0.125 \\
 & Few-shot & 0.687 (0.665 - 0.709) & 0.223 (0.199 - 0.251) & 0.074 & 0.605 (0.585 - 0.626) & 0.261 (0.241 - 0.286) & 0.183 \\
 & CoT & 0.640 (0.617 - 0.666) & 0.202 (0.180 - 0.228) & 0.059 & 0.568 (0.548 - 0.587) & 0.224 (0.208 - 0.244) & 0.220 \\
 & CoT-SC & 0.659 (0.637 - 0.683) & 0.221 (0.197 - 0.249) & 0.086 & 0.592 (0.570 - 0.612) & 0.253 (0.234 - 0.279) & 0.239 \\
 & Self-Refine & 0.630 (0.606 - 0.654) & 0.189 (0.169 - 0.216) & 0.065 & 0.568 (0.548 - 0.586) & 0.224 (0.207 - 0.244) & 0.151 \\
\midrule
\multirow{5}{*}{\textbf{HuaTuo}}  & Zero-shot & 0.692 (0.672 - 0.714) & \textbf{0.238 (0.213 - 0.268)} & 0.166 & 0.599 (0.578 - 0.620) & 0.256 (0.236 - 0.281) & 0.203 \\
 & Few-shot & \textbf{0.700 (0.681 - 0.721)} & 0.227 (0.205 - 0.258) & 0.093 & 0.602 (0.580 - 0.621) & 0.260 (0.240 - 0.284) & 0.029 \\
 & CoT & 0.670 (0.650 - 0.693) & 0.219 (0.194 - 0.247) & 0.211 & 0.595 (0.574 - 0.615) & 0.258 (0.239 - 0.280) & 0.599 \\
 & CoT-SC & 0.670 (0.649 - 0.692) & 0.218 (0.194 - 0.246) & 0.211 & 0.594 (0.574 - 0.614) & 0.258 (0.239 - 0.281) & 0.599 \\
 & Self-Refine & 0.671 (0.651 - 0.694) & 0.213 (0.190 - 0.242) & 0.106 & 0.581 (0.560 - 0.601) & 0.249 (0.230 - 0.273) & 0.385 \\
\midrule
\multirow{5}{*}{\textbf{Llava}}  & Zero-shot & 0.642 (0.621 - 0.666) & 0.204 (0.182 - 0.235) & 0.755 & 0.574 (0.552 - 0.594) & 0.234 (0.216 - 0.255) & 0.787 \\
 & Few-shot & 0.684 (0.664 - 0.706) & 0.233 (0.208 - 0.265) & 0.776 & 0.605 (0.584 - 0.624) & 0.253 (0.235 - 0.273) & 0.780 \\
 & CoT & 0.633 (0.610 - 0.655) & 0.170 (0.154 - 0.186) & 0.831 & 0.509 (0.499 - 0.518) & 0.194 (0.183 - 0.205) & 0.808 \\
 & CoT-SC & 0.645 (0.622 - 0.666) & 0.197 (0.174 - 0.221) & 0.840 & 0.542 (0.523 - 0.560) & 0.206 (0.194 - 0.221) & 0.806 \\
 & Self-Refine & 0.631 (0.608 - 0.655) & 0.170 (0.154 - 0.187) & 0.829 & 0.516 (0.504 - 0.526) & 0.196 (0.185 - 0.208) & 0.808 \\
\bottomrule
\end{tabular}}
\end{table*}

\subsection{Implementation Details}
To ensure a rigorous evaluation, we instantiate all agentic baselines using four distinct VLM backbones, two general-purpose and two specialized medical. All models utilize approximately 7-8 billion parameters to maintain comparable computational requirements.

\subsubsection{Model Backbones}
\begin{enumerate}
    \item \textbf{Qwen2.5-VL-7B-Instruct} \citep{bai_qwen25-vl_2025}: A leading generalist VLM built upon the Qwen2.5 language model. It features a specialized vision encoder optimized for high-resolution image understanding, making it a strong generalist baseline.
    
    \item \textbf{InternVL2.5-8B-MPO} \citep{chen_expanding_2025}: A general-purpose VLM from OpenGVLab designed for robust visual reasoning. This variant employs Mixed Preference Optimization (MPO) to align the model's outputs with human preference.
    
    \item \textbf{HuatuoGPT-Vision-7B-Qwen2.5VL} \citep{chen_huatuogpt-vision_2024}: A specialized medical VLM initialized from Qwen2.5-VL. It is further post-trained on a massive corpus of multimodal medical data.
    
    \item \textbf{LLaVA-Med-v1.5-Mistral-7B} \citep{li_llava-med_2023}: A biomedical VLM that builds on the Mistral-7B language model. It was pre-trained on a large-scale dataset of biomedical figure-caption pairs (PMC-15M).
\end{enumerate}

\subsubsection{Hyperparameters}
All experiments utilized a standard batch size of 3 or 4, with exceptions for the supervised architectures (batch size 16) and multimodal LLaVA-Med, which required a batch size of 1 to accommodate heterogeneous modality combinations. Regarding context constraints, the high token overhead of serialized tabular EHR data exceeded the effective limit of InternVL2.5, necessitating its evaluation on a restricted 3-modality subset (PS, CXR, RR) in Appendix \ref{tab:intern_multimodal}.

 %We ran a random search for optimizing the learning rates for mortality, and use the best learning rates for both tasks.

\begin{table*}[ht]
\caption{\textbf{Multimodal Results.} Comparison of Supervised, Single-Agent, and Multi-Agent architectures on the 4-modality dataset. Best overall performance in each column is bolded.}
\label{tab:main_multimodal}
\centering
\resizebox{\linewidth}{!}{%
\begin{tabular}{lllcccccc}
\toprule
\multirow{2}{*}{\textbf{Backbone}} & \multirow{2}{*}{\textbf{Arch.}} & \multirow{2}{*}{\textbf{Method}} & \multicolumn{3}{c}{\textbf{In-Hospital Mortality}} & \multicolumn{3}{c}{\textbf{Length of Stay ($>$7 Days)}} \\
\cmidrule(lr){4-6} \cmidrule(lr){7-9}
 & & & \textbf{AUROC} & \textbf{AUPRC} & \textbf{ECE} & \textbf{AUROC} & \textbf{AUPRC} & \textbf{ECE} \\
\midrule
Supervised & - & MedPatch & \textbf{0.877 (0.864 - 0.888)} & \textbf{0.546 (0.504 - 0.585)} & \textbf{0.019} & \textbf{0.844 (0.830 - 0.857)} & \textbf{0.551 (0.517 - 0.587)} & \textbf{0.025} \\
\midrule
\multirow{8}{*}{\textbf{Qwen}} & \multirow{4}{*}{Single} 
 & Zero-shot & 0.756 (0.737 - 0.776) & 0.330 (0.297 - 0.368) & 0.023 & 0.714 (0.694 - 0.731) & 0.345 (0.320 - 0.373) & 0.411 \\
 & & Few-shot & 0.763 (0.744 - 0.783) & 0.325 (0.292 - 0.361) & 0.032 & 0.682 (0.662 - 0.699) & 0.318 (0.294 - 0.346) & 0.479 \\
 & & CoT & 0.733 (0.713 - 0.752) & 0.274 (0.244 - 0.308) & 0.049 & 0.683 (0.663 - 0.702) & 0.311 (0.288 - 0.336) & 0.676 \\
 & & CoT-SC & 0.762 (0.742 - 0.782) & 0.337 (0.300 - 0.379) & 0.039 & 0.698 (0.678 - 0.715) & 0.340 (0.313 - 0.370) & 0.661 \\
 \cmidrule{2-9}
 & \multirow{4}{*}{Multi} 
 & Majority Vote & 0.748 (0.727 - 0.768) & 0.315 (0.282 - 0.355) & 0.111 & 0.710 (0.690 - 0.728) & 0.352 (0.324 - 0.383) & 0.046 \\
 & & Debate & 0.631 (0.607 - 0.656) & 0.210 (0.185 - 0.243) & 0.091 & 0.644 (0.623 - 0.664) & 0.281 (0.259 - 0.307) & 0.121 \\
 & & Meta-Prompt & 0.599 (0.573 - 0.625) & 0.179 (0.160 - 0.204) & 0.051 & 0.537 (0.514 - 0.558) & 0.226 (0.208 - 0.249) & 0.086 \\
 & & Traj-CoA & 0.762 (0.743 - 0.780) & 0.318 (0.283 - 0.355) & 0.039 & 0.708 (0.688 - 0.726) & 0.336 (0.310 - 0.365) & 0.190 \\
  & & MDAgents & 0.624 (0.601 - 0.647) & 0.192 (0.169 - 0.221) & 0.110 & 0.584 (0.563 - 0.603) & 0.240 (0.222 - 0.262) & 0.138 \\
 & & MedAgents & 0.662 (0.641 - 0.686) & 0.206 (0.185 - 0.235) & 0.034 & 0.634 (0.613 - 0.654) & 0.285 (0.261 - 0.310) & 0.019 \\
\midrule
\multirow{8}{*}{\textbf{HuaTuo}} & \multirow{4}{*}{Single} 
 & Zero-shot & 0.762 (0.743 - 0.782) & 0.325 (0.296 - 0.362) & 0.049 & 0.704 (0.686 - 0.721) & 0.331 (0.307 - 0.358) & 0.418 \\
 & & Few-shot & 0.763 (0.744 - 0.782) & 0.324 (0.291 - 0.361) & 0.167 & 0.703 (0.684 - 0.720) & 0.332 (0.308 - 0.358) & 0.514 \\
 & & CoT & 0.697 (0.678 - 0.720) & 0.233 (0.208 - 0.266) & 0.306 & 0.639 (0.617 - 0.658) & 0.282 (0.260 - 0.306) & 0.671 \\
 & & CoT-SC & 0.696 (0.678 - 0.719) & 0.233 (0.208 - 0.265) & 0.305 & 0.639 (0.617 - 0.658) & 0.282 (0.261 - 0.306) & 0.669 \\
 \cmidrule{2-9}
 & \multirow{4}{*}{Multi} 
 & Majority Vote & 0.711 (0.690 - 0.730) & 0.245 (0.217 - 0.279) & 0.050 & 0.691 (0.671 - 0.709) & 0.322 (0.296 - 0.352) & 0.087 \\
 & & Debate & 0.628 (0.604 - 0.652) & 0.188 (0.165 - 0.216) & 0.109 & 0.621 (0.599 - 0.642) & 0.273 (0.250 - 0.300) & 0.061 \\
 & & Meta-Prompt & 0.636 (0.614 - 0.657) & 0.179 (0.160 - 0.202) & 0.428 & 0.558 (0.534 - 0.578) & 0.228 (0.210 - 0.250) & 0.080 \\
 & & Traj-CoA & 0.744 (0.725 - 0.766) & 0.295 (0.266 - 0.333) & 0.065 & 0.701 (0.680 - 0.718) & 0.326 (0.301 - 0.353) & 0.323 \\
 & & MDAgents & 0.490 (0.467 - 0.514) & 0.125 (0.112 - 0.142) & 0.309 & 0.571 (0.551 - 0.590)) & 0.237 (0.218 - 0.260) & 0.059 \\
 & & MedAgents & 0.606 (0.584 - 0.629) & 0.165 (0.148 - 0.187) & 0.335 & 0.584 (0.563 - 0.606) & 0.230 (0.213 - 0.252) & 0.486 \\
\midrule
\multirow{8}{*}{\textbf{Llava}} & \multirow{4}{*}{Single} 
 & Zero-shot & 0.741 (0.721 - 0.761) & 0.268 (0.242 - 0.303) & 0.835 & 0.613 (0.596 - 0.630) & 0.242 (0.226 - 0.259) & 0.803 \\
 & & Few-shot & 0.684 (0.662 - 0.706) & 0.225 (0.199 - 0.255) & 0.704 & 0.581 (0.561 - 0.602) & 0.233 (0.217 - 0.252) & 0.786 \\
 & & CoT & 0.676 (0.658 - 0.693) & 0.184 (0.167 - 0.202) & 0.819 & 0.553 (0.537 - 0.567) & 0.210 (0.197 - 0.223) & 0.807 \\
 & & CoT-SC & 0.691 (0.670 - 0.714) & 0.215 (0.192 - 0.240) & 0.806 & 0.583 (0.563 - 0.602) & 0.226 (0.211 - 0.242) & 0.806 \\
 \cmidrule{2-9}
 & \multirow{4}{*}{Multi} 
 & Majority Vote & 0.710 (0.686 - 0.732) & 0.280 (0.248 - 0.316) & 0.812 & 0.674 (0.654 - 0.693) & 0.291 (0.268 - 0.316) & 0.765 \\
 & & Debate & 0.495 (0.470 - 0.519) & 0.121 (0.109 - 0.137) & 0.293 & 0.506 (0.485 - 0.525) & 0.192 (0.178 - 0.210) & 0.437 \\
 & & Meta-Prompt & 0.659 (0.638 - 0.683) & 0.211 (0.189 - 0.237) & 0.810 & 0.576 (0.555 - 0.594) & 0.231 (0.214 - 0.250) & 0.780 \\
 & & Traj-CoA & 0.694 (0.673 - 0.718) & 0.249 (0.221 - 0.283) & 0.758 & 0.608 (0.589 - 0.626) & 0.239 (0.223 - 0.257) & 0.795 \\
 & & MDAgents & 0.565 (0.544 - 0.588) & 0.143 (0.130 - 0.160) & 0.771 & 0.492 (0.473 - 0.512)  & 0.194 (0.181 - 0.210) & 0.759 \\
 & & MedAgents & 0.609 (0.589 - 0.633) & 0.164 (0.147 - 0.184) & 0.683 & 0.545 (0.525 - 0.563) & 0.207 (0.193 - 0.223) & 0.782 \\
\bottomrule
\end{tabular}}
\end{table*}

\section{Results}
%In this section, we present our main experimental findings on in-hospital mortality and long length of stay prediction, including unimodal and multimodal performance results as well as ablation experiments that highlight the contributions of incrementally adding modalities in single agent and multi-agent settings.

\subsection{Unimodal Results}
In Table \ref{tab:unimodal}, we present the Single Agent Unimodal setup where agents are restricted to the PS modality. %This setting allows us to evaluate the base reasoning and medical knowledge retrieval capabilities of the agents.
%\paragraph{In-Hospital Mortality Task}
For the mortality prediction task, specialized LLMs demonstrate the capacity to exceed supervised baselines. While BioBERT achieves a competitive AUROC of $0.680$, the medical-specific backbone HuaTuo surpasses this in the few-shot setting with an AUROC of $0.700$. In terms of precision-recall balance, HuaTuo Zero-shot attains the highest AUPRC of $0.238$, improving upon the BioBERT baseline of $0.228$. Generalist models also show competence with Qwen few-shot achieving an AUROC of $0.697$, closely following the specialized medical agents. 

Despite strong discriminative performance (AUROC/AUPRC), generative backbones exhibit significant miscalibration compared to discriminative models. The supervised BioBERT baseline maintains a minimal ECE of $0.006$. Conversely, all agentic setups yield considerably higher ECE values. For instance, the top-performing HuaTuo Few-shot achieves an ECE of $0.093$, while Llava configurations consistently exceed $0.750$. This indicates that while agents can rank patient risk effectively, their probabilistic confidence scores are less reliable than the supervised baseline.

%\paragraph{Length of Stay (LOS) Task}
For LoS prediction, BioBERT achieves the highest performance across all metrics with an AUROC of $0.641$ and AUPRC of $0.307$. Among the agentic backbones, the generalist Intern model performs best, with the zero-shot setup achieving an AUROC of $0.620$ and AUPRC of $0.278$. Notably, the medical-specific HuaTuo backbone underperforms in this task (zero-shot AUROC $0.599$), suggesting that clinical pre-training may be more beneficial for diagnostic tasks like mortality than for operational metrics like LOS.

%\paragraph{Impact of Prompting Strategies}
% Analysis of prompting strategies reveals that increased complexity does not consistently yield performance gains in the unimodal setting. Across backbones, zero-shot and few-shot configurations generally outperform CoT and SR frameworks. For example, for in-hospital mortality, Qwen CoT yields an AUROC of $0.650$, inferior to both its zero-shot ($0.667$) and few-shot ($0.697$) counterparts. Similarly, Intern CoT ($0.660$) shows no improvement over Intern zero-shot ($0.663$). This trend suggests that for single-modality tasks, the added noise or token length from complex reasoning chains may impede classification accuracy.

\subsection{Multimodal Results}
Table \ref{tab:main_multimodal} presents the performance of agentic backbones when processing multimodal clinical data. A performance gap persists between generalist agents and the supervised MedPatch for both tasks. MedPatch achieves an AUROC of $0.877$ for in-hospital mortality and $0.844$ for length of stay, whereas the best-performing agent configurations trail significantly behind at $0.763$ (HuaTuo Few-shot)  and $0.714$ (Qwen Zero-shot). This disparity highlights a structural limitation: while LLMs possess strong semantic reasoning capabilities, they lack the dedicated fusion layers found in specialized architectures optimized for multimodal inputs. Despite the gap however, integrating multiple modalities still yields considerable improvements over unimodal baselines in the single agent settings. For the Qwen backbone, mortality prediction performance improves from a unimodal baseline of $0.667$ to $0.756$ in the multimodal zero-shot setting. Likewise, LOS performance for Qwen increases from $0.603$ to $0.714$.

% , despite utilizing only three modalities, demonstrates competitive reasoning, achieving a peak mortality AUROC of $0.735$ in the Few-shot setting and an LOS AUROC of $0.731$ using the Traj-CoA architecture.

%\paragraph{Inference Strategy Variability}
% The efficacy of inference strategies is backbone- and task-dependent. Self-Consistency (CoT-SC) yields the optimal mortality prediction for Qwen, achieving an AUROC of $0.773$ compared to $0.756$ for Zero-shot. However, this benefit does not generalize. For HuaTuo, CoT-SC ($0.696$) performs worse than the Zero-shot baseline ($0.762$). Furthermore, complex prompting strategies often degrade LOS performance. For instance, Qwen CoT ($0.683$) and CoT-SC ($0.698$) both trail the simpler Zero-shot setup ($0.714$).

%\paragraph{Multi-Agent vs. Single-Agent Architectures}
We test the hypothesis that distributing reasoning across specialized agents improves outcomes. Collaborative protocols often result in performance degradation. In Table \ref{tab:main_multimodal}, Qwen-based Debate and Meta-Prompt architectures yield mortality AUROCs of $0.631$ and $0.599$, respectively, lower than the single agent zero-shot baseline ($0.756$). Among multi-agent frameworks, Traj-CoA consistently performs best (e.g., Qwen Mortality $0.762$). Unlike fully decentralized baselines, this architecture employs a final decision-maker that retains direct access to raw multimodal data while incorporating the intermediate reasoning trajectories of the specialized EHR agents.

\begin{table*}[t!]
\caption{\textbf{Modality Ablations.} Comparing zero-shot vs. majority vote settings for in-hospital mortality. Note that while AUROC/AUPRC scale with modality density, Single Agent calibration (ECE) improves with more data while majority vote calibration deteriorates. Best overall performance in each column is bolded.}
\label{tab:combined_ablations}
\centering
\resizebox{\linewidth}{!}{%
\begin{tabular}{lllccc}
\toprule
\multirow{2}{*}{\textbf{Backbone}} & \multirow{2}{*}{\textbf{Arch.}} & \multirow{2}{*}{\textbf{Modalities}} & \multicolumn{3}{c}{\textbf{In-Hospital Mortality}} \\
\cmidrule(lr){4-6}
 & & & \textbf{AUROC} & \textbf{AUPRC} & \textbf{ECE} \\
\midrule
\multirow{7}{*}{\textbf{Qwen}} & \multirow{4}{*}{Single (ZS)} 
  & PS & 0.667 (0.645 - 0.690) & 0.234 (0.208 - 0.267) & 0.025 \\
 & & PS + CXR & 0.671 (0.649 - 0.695) & 0.236 (0.209 - 0.267) & 0.027 \\
 & & PS + CXR + RR & 0.732 (0.712 - 0.753) & 0.273 (0.244 - 0.305) & 0.027 \\
 & & All (PS+EHR+CXR+RR) & 0.756 (0.737 - 0.776) & \textbf{0.330 (0.297 - 0.368)} & 0.023 \\
 \cmidrule{2-6}
 & \multirow{3}{*}{Multi (MV)} 
  & PS + CXR & 0.666 (0.643 - 0.689) & 0.221 (0.197 - 0.252) & 0.068 \\
 & & PS + CXR + RR & 0.725 (0.704 - 0.745) & 0.266 (0.237 - 0.301) & 0.089 \\
 & & All (PS+EHR+CXR+RR) & 0.748 (0.727 - 0.768) & 0.315 (0.282 - 0.355) & 0.111 \\
\midrule
\multirow{7}{*}{\textbf{HuaTuo}} & \multirow{4}{*}{Single (ZS)} 
  & PS & 0.692 (0.672 - 0.714) & 0.238 (0.213 - 0.268) & 0.166 \\
 & & PS + CXR & 0.695 (0.675 - 0.716) & 0.236 (0.212 - 0.266) & 0.205 \\
 & & PS + CXR + RR & 0.742 (0.723 - 0.763) & 0.277 (0.248 - 0.310) & 0.269 \\
 & & All (PS+EHR+CXR+RR) & \textbf{0.762 (0.743 - 0.782)} & 0.325 (0.296 - 0.362) & 0.049 \\
 \cmidrule{2-6}
 & \multirow{3}{*}{Multi (MV)} 
  & PS + CXR & 0.649 (0.629 - 0.670) & 0.201 (0.177 - 0.229) & 0.068 \\
 & & PS + CXR + RR & 0.689 (0.667 - 0.709) & 0.220 (0.195 - 0.252) & \textbf{0.014} \\
 & & All (PS+EHR+CXR+RR) & 0.711 (0.690 - 0.730) & 0.245 (0.217 - 0.279) & 0.050 \\
\midrule
\multirow{7}{*}{\textbf{Llava}} & \multirow{4}{*}{Single (ZS)} 
  & PS & 0.642 (0.621 - 0.666) & 0.204 (0.182 - 0.235) & 0.755 \\
 & & PS + CXR & 0.658 (0.636 - 0.681) & 0.213 (0.188 - 0.243) & 0.766 \\
 & & PS + CXR + RR & 0.734 (0.712 - 0.756) & 0.270 (0.241 - 0.304) & 0.779 \\
 & & All (PS+EHR+CXR+RR) & 0.741 (0.721 - 0.761) & 0.268 (0.242 - 0.303) & 0.835 \\
 \cmidrule{2-6}
 & \multirow{3}{*}{Multi (MV)} 
  & PS + CXR & 0.621 (0.597 - 0.647) & 0.184 (0.164 - 0.211) & 0.804 \\
 & & PS + CXR + RR & 0.694 (0.672 - 0.717) & 0.256 (0.227 - 0.291) & 0.800 \\
 & & All (PS+EHR+CXR+RR) & 0.711 (0.690 - 0.730) & 0.245 (0.217 - 0.279) & 0.812 \\
\bottomrule
\end{tabular}}
\end{table*}

\begin{table}
\centering
\caption{Consensus analysis showing the percentage of samples reaching consensus at each round and AUROC.}
\label{tab:consensus}
\resizebox{\linewidth}{!}{%
\begin{tabular}{lccccc}
\toprule
Backbone & Round 1 & Round 2 & Round 3 & MAX & AUROC \\
\midrule
LlaVaMed & 47.1\% & 0.2\% & 0.2\% & 52.6\% & 0.495 \\
Huatuo   & 96.8\% & 2.1\% & 0.3\% & 0.8\%  & 0.628 \\
Qwen     & 100\%  & 0\%   & 0\%   & 0\%    & 0.631 \\
\bottomrule
\end{tabular}}
\end{table}

The supervised MedPatch baseline maintains a low ECE of $0.0189$. While the Qwen single agent zero-shot setup remains relatively calibrated (ECE $0.0233$), the majority vote architecture exhibits an increase in calibration error to $0.111$. This indicates that voting mechanisms may distort the probabilistic confidence of the system. Intern Traj-CoA defies this trend in specific instances, achieving an ECE of $0.010$ for mortality prediction. Another observation is that different backbone have varying capabilities when it comes to handling multimodal data. This is evident in Appendix \ref{calibration} where different backbones show different reductions in ECE with added modalities. Some like Intern show an increase in ECE with more modalities added.

\subsection{Ablations}
Table \ref{tab:combined_ablations} presents an ablation study for the in-hospital mortality task, revealing that single agent architectures benefit significantly from increased modality integration. As predictive performance scales positively with the addition of modalities, calibration also improves, with the Qwen backbone seeing AUROC rise from $0.667$ to $0.756$ while its ECE drops to $0.0233$. A similar trend appears for HuaTuo and LLava. This supports the idea that single agents effectively leverage cross-modal dependencies within heterogeneous data to enhance performance.

In contrast, majority vote demonstrates a divergence in reliability, exhibiting degrading calibration as the system complexity increases. Despite the contribution of specialized agents to classification accuracy, the aggregation of hard labels without shared context leads to significant system overconfidence, illustrated by Qwen's ECE rising from $0.068$ to $0.111$ in the full multimodal setting. This suggests that decentralized voting mechanisms fail to capture the uncertainty reduction benefits inherent in multimodal data, unlike the unified processing found in single-agent setups. The ablation results are visualized in Appendix \ref{Figure3}. Considering the different performance improvements noticed as modalities are added to the majority vote, we experiment with a weighted majority vote baseline in Appendix \ref{tab:WMV_performance}. In that baseline, we weigh each modality agent by its unimodal AUROC and notice minor performance improvements.

To further understand the failure modes of multi-agent systems, we analyze the generated traces and conduct more ablations. Table \ref{tab:consensus} shows the percentage of samples in the debate baseline traces that reached consensus at each round, or never reached consensus and had their probabilities averaged (MAX rounds). The results show a big variety depending on the backbone used. For instance, the Qwen backbone reached consensus for all the samples from the first round (indicating no inter-agent debate), while LLavaMed never reached a consensus for half of the patients. Notably, the models show a trend where having more samples with more rounds of debate correlates with degraded performance in AUROC. This suggests that current debate capabilities are lacking and do not consistently improve agentic reasoning. Beyond consensus, Appendix \ref{tab:sycophancy} summarizes the echo chamber behavior observed across the backbones when used for the debate baseline. We also experiment with a debate between multimodal agents (in Appendix \ref{tab:debate_multimodal}), where 4 full-modality agents undergo 3 rounds of debate until consensus. We note that the performance there still lags behind the single agent baselines. Considering the trends in consensus, sycophancy, and the multimodal debate results, it seems that multi-agent systems suffer from a combination of information loss, noise propagation, and weak modality fusion. This is further evidenced by the Traj-CoA baseline's performance, where the presence of a multimodal decision-making agent, closely resembling the single-agent setup, reduces the observed errors.

% Acknowledgements should only appear in the accepted version.
% \section*{Acknowledgements}

% \textbf{Do not} include acknowledgements in the initial version of the paper
% submitted for blind review.

% If a paper is accepted, the final camera-ready version can (and usually should)
% include acknowledgements.  Such acknowledgements should be placed at the end of
% the section, in an unnumbered section that does not count towards the paper
% page limit. Typically, this will include thanks to reviewers who gave useful
% comments, to colleagues who contributed to the ideas, and to funding agencies
% and corporate sponsors that provided financial support.

\section{Discussion}
\label{sec:discussion}

In this work, we introduced \texttt{AgentRx}, a first-of-its-kind benchmark for evaluating LLM-based agentic systems on high-stakes clinical risk prediction tasks. We systematically analyzed how different agentic architectures, ranging from single-agent baselines to complex multi-agent frameworks, perform against supervised deep learning models. The study yields a critical insight that multi-agent multimodal systems consistently underperform single-agent multimodal systems. This is driven by a fundamental divergence in system calibration where we observe a striking contrast in how single agent systems improve calibration with additional data modalities while multi-agent systems degrade. Closing this calibration gap is vital for the development of scalable decentralized healthcare systems.

More generally, our findings align with the recent benchmarks established by \cite{zhu_medagentboard_2025} in \texttt{MedAgentBoard}, confirming that agentic systems generally trail state-of-the-art supervised fusion networks in clinical prediction. However, we identify a critical modality-dependent nuance to this observation. The analysis in \texttt{MedAgentBoard} largely focuses on structured EHR data, where LLMs struggle to encode high-dimensional numerical features. In contrast, our unimodal setting utilizing PS which is a free-text modality demonstrates that specialized agents can effectively outperform supervised baselines. This suggests that the performance drop is not intrinsic to the task of risk prediction, but rather a consequence of the modality. Notably, LLM agents seem to synthesize multimodal data more effectively when the task at hand is diagnostic in nature. This is shown in Appendix \ref{tab:qwen_ckd}, where both the single agent and multi-agent systems outperform the supervised baselines in the multimodal setting on the note-based chronic kidney disease detection task.
% Our architectural analysis further highlights the trade-off between decentralized specialization and holistic synthesis. While \citet{zhu_medagentboard_2025} observed inconsistent gains with multi-agent collaboration, our results in the multimodal setting are far more definitive: decentralized architectures consistently underperformed. This suggests that without a centralized mechanism to ground reasoning (as seen in our Traj-CoA setup), the added complexity of multi-agent communication introduces noise rather than signal.

Despite these contributions, this study has limitations. Our benchmark relies on the MIMIC database, which represents a single-center cohort and may pose challenges for generalizability. Additionally, we have limited our evaluation to architectures of sizes within 7-8 billion parameters. We acknowledge that smaller or larger models may exhibit different reasoning capabilities. Appendix \ref{tab:medgemma} summarizes the performance of a smaller backbone family on some frameworks from our benchmark. Finally, the serialization of high-frequency EHR data into text-based context windows remains an open research problem that likely limits the agents' ability to capture subtle physiological trends. In Appendix \ref{tab:log_comparison}, we explore some alternative serialization strategies. Moving forward, we plan to expand \texttt{AgentRx} to include more diverse clinical endpoints and investigate architectures that combine the interpretability of LLM reasoning with the predictive precision of frozen, supervised encoders. Given the current limitations of the examined baselines in handling complex multimodal data, future work could focus on moving away from purely textual communication and exploring latent-based representations or ICU-specific tool-calling frameworks to enhance reasoning and risk prediction.

\section*{Acknowledgements}
This work was supported by ASPIRE, the technology program management pillar of Abu Dhabi’s Advanced Technology Research Council (ATRC), via the ASPIRE Precision Medicine Research Institute Abu Dhabi (ASPIREPMRIAD) award grant number VRI-20-10, the NYUAD Center for Artificial Intelligence and Robotics, funded by Tamkeen under the NYUAD Research Institute Award CG010, the Meem Foundation, and the ADIA Lab Health Sciences Grant. The research was carried out on the High Performance Computing resources at New York University Abu Dhabi. Figure \ref{Figure1} was created in BioRender (Al Jorf, B. (2025) https://BioRender.com/jl2gr7s).

\bibliography{ref}

\begin{thebibliography}{54}
\providecommand{\natexlab}[1]{#1}
\providecommand{\url}[1]{\texttt{#1}}
\expandafter\ifx\csname urlstyle\endcsname\relax
  \providecommand{\doi}[1]{doi: #1}\else
  \providecommand{\doi}{doi: \begingroup \urlstyle{rm}\Url}\fi

\bibitem[Acharya et~al.(2024)Acharya, Shrestha, Chen, Conte, Avramovic, Sikdar, Anastasopoulos, and Das]{acharya_clinical_2024}
Angeela Acharya, Sulabh Shrestha, Anyi Chen, Joseph Conte, Sanja Avramovic, Siddhartha Sikdar, Antonios Anastasopoulos, and Sanmay Das.
\newblock Clinical risk prediction using language models: benefits and considerations.
\newblock \emph{Journal of the American Medical Informatics Association: JAMIA}, 31\penalty0 (9):\penalty0 1856--1864, September 2024.
\newblock ISSN 1527-974X.
\newblock \doi{10.1093/jamia/ocae030}.

\bibitem[Afshar et~al.(2025)Afshar, Ryan~Baumann, Resnik, Hintzke, Gravel~Sullivan, Wills, Lemmon, Dambach, Mrotek, Quinn, Abramson, Kleinschmidt, Brazelton, Leaf, Twedt, Kunstman, Patterson, Liao, Rasmussen, Burnside, Goswami, and Gordon]{afshar_pragmatic_2025}
Majid Afshar, Mary Ryan~Baumann, Felice Resnik, Josie Hintzke, Anne Gravel~Sullivan, Graham Wills, Kayla Lemmon, Jason Dambach, Leigh~Ann Mrotek, Mariah Quinn, Kirsten Abramson, Peter Kleinschmidt, Thomas~B. Brazelton, Margaret~A. Leaf, Heidi Twedt, David Kunstman, Brian Patterson, Frank Liao, Stacy Rasmussen, Elizabeth~S. Burnside, Cherodeep Goswami, and Joel Gordon.
\newblock A {Pragmatic} {Randomized} {Controlled} {Trial} of {Ambient} {Artificial} {Intelligence} to {Improve} {Health} {Practitioner} {Well}-{Being}.
\newblock \emph{NEJM AI}, 2\penalty0 (12):\penalty0 AIoa2500945, November 2025.
\newblock \doi{10.1056/AIoa2500945}.
\newblock URL \url{https://ai.nejm.org/doi/10.1056/AIoa2500945}.
\newblock Publisher: Massachusetts Medical Society.

\bibitem[Al~Jorf et~al.(2026)Al~Jorf, Piechowski-Jozwiak, and Shamout]{al_jorf_data-centric_2026}
Baraa Al~Jorf, Bartlomiej Piechowski-Jozwiak, and Farah~E. Shamout.
\newblock A data-centric perspective on designing {AI} foundation models for healthcare.
\newblock \emph{Frontiers in Digital Health}, Volume 8 - 2026, 2026.
\newblock ISSN 2673-253X.
\newblock \doi{10.3389/fdgth.2026.1738523}.
\newblock URL \url{https://www.frontiersin.org/journals/digital-health/articles/10.3389/fdgth.2026.1738523}.

\bibitem[Bae et~al.(2023)Bae, Kyung, Ryu, Cho, Lee, Kweon, Oh, Ji, Chang, Kim, and Choi]{bae_ehrxqa_2023}
Seongsu Bae, Daeun Kyung, Jaehee Ryu, Eunbyeol Cho, Gyubok Lee, Sunjun Kweon, Jungwoo Oh, Lei Ji, Eric Chang, Tackeun Kim, and Edward Choi.
\newblock {EHRXQA}: {A} {Multi}-{Modal} {Question} {Answering} {Dataset} for {Electronic} {Health} {Records} with {Chest} {X}-ray {Images}.
\newblock \emph{Advances in Neural Information Processing Systems}, 36:\penalty0 3867--3880, December 2023.
\newblock URL \url{https://proceedings.neurips.cc/paper_files/paper/2023/hash/0c007ebef1d11fd48da6ce4f54687db6-Abstract-Datasets_and_Benchmarks.html}.

\bibitem[Bai et~al.(2025)Bai, Chen, Liu, Wang, Ge, Song, Dang, Wang, Wang, Tang, Zhong, Zhu, Yang, Li, Wan, Wang, Ding, Fu, Xu, Ye, Zhang, Xie, Cheng, Zhang, Yang, Xu, and Lin]{bai_qwen25-vl_2025}
Shuai Bai, Keqin Chen, Xuejing Liu, Jialin Wang, Wenbin Ge, Sibo Song, Kai Dang, Peng Wang, Shijie Wang, Jun Tang, Humen Zhong, Yuanzhi Zhu, Mingkun Yang, Zhaohai Li, Jianqiang Wan, Pengfei Wang, Wei Ding, Zheren Fu, Yiheng Xu, Jiabo Ye, Xi~Zhang, Tianbao Xie, Zesen Cheng, Hang Zhang, Zhibo Yang, Haiyang Xu, and Junyang Lin.
\newblock Qwen2.5-{VL} {Technical} {Report}, February 2025.
\newblock URL \url{http://arxiv.org/abs/2502.13923}.
\newblock arXiv:2502.13923 [cs].

\bibitem[Bicknell et~al.(2024)Bicknell, Butler, Whalen, Ricks, Dixon, Clark, Spaedy, Skelton, Edupuganti, Dzubinski, Tate, Dyess, Lindeman, and Lehmann]{bicknell_chatgpt-4_2024}
Brenton~T. Bicknell, Danner Butler, Sydney Whalen, James Ricks, Cory~J. Dixon, Abigail~B. Clark, Olivia Spaedy, Adam Skelton, Neel Edupuganti, Lance Dzubinski, Hudson Tate, Garrett Dyess, Brenessa Lindeman, and Lisa~Soleymani Lehmann.
\newblock {ChatGPT}-4 {Omni} {Performance} in {USMLE} {Disciplines} and {Clinical} {Skills}: {Comparative} {Analysis}.
\newblock \emph{JMIR Medical Education}, 10\penalty0 (1):\penalty0 e63430, November 2024.
\newblock \doi{10.2196/63430}.
\newblock URL \url{https://mededu.jmir.org/2024/1/e63430}.

\bibitem[Brown et~al.(2020)Brown, Mann, Ryder, Subbiah, Kaplan, Dhariwal, Neelakantan, Shyam, Sastry, Askell, Agarwal, Herbert-Voss, Krueger, Henighan, Child, Ramesh, Ziegler, Wu, Winter, Hesse, Chen, Sigler, Litwin, Gray, Chess, Clark, Berner, McCandlish, Radford, Sutskever, and Amodei]{brown_language_2020}
Tom~B. Brown, Benjamin Mann, Nick Ryder, Melanie Subbiah, Jared Kaplan, Prafulla Dhariwal, Arvind Neelakantan, Pranav Shyam, Girish Sastry, Amanda Askell, Sandhini Agarwal, Ariel Herbert-Voss, Gretchen Krueger, Tom Henighan, Rewon Child, Aditya Ramesh, Daniel~M. Ziegler, Jeffrey Wu, Clemens Winter, Christopher Hesse, Mark Chen, Eric Sigler, Mateusz Litwin, Scott Gray, Benjamin Chess, Jack Clark, Christopher Berner, Sam McCandlish, Alec Radford, Ilya Sutskever, and Dario Amodei.
\newblock Language {Models} are {Few}-{Shot} {Learners}, May 2020.
\newblock URL \url{https://arxiv.org/abs/2005.14165v4}.

\bibitem[Catalina et~al.(2023)Catalina, Fuster-Casanovas, Vidal-Alaball, Escalé-Besa, Marin-Gomez, Femenia, and Solé-Casals]{catalina_knowledge_2023}
Queralt~Miró Catalina, Aïna Fuster-Casanovas, Josep Vidal-Alaball, Anna Escalé-Besa, Francesc~X Marin-Gomez, Joaquim Femenia, and Jordi Solé-Casals.
\newblock Knowledge and perception of primary care healthcare professionals on the use of artificial intelligence as a healthcare tool.
\newblock \emph{DIGITAL HEALTH}, 9:\penalty0 20552076231180511, January 2023.
\newblock ISSN 2055-2076.
\newblock \doi{10.1177/20552076231180511}.
\newblock URL \url{https://doi.org/10.1177/20552076231180511}.
\newblock Publisher: SAGE Publications Ltd.

\bibitem[Cemri et~al.(2025)Cemri, Pan, Yang, Agrawal, Chopra, Tiwari, Keutzer, Parameswaran, Klein, Ramchandran, Zaharia, Gonzalez, and Stoica]{cemri_why_2025}
Mert Cemri, Melissa~Z. Pan, Shuyi Yang, Lakshya~A. Agrawal, Bhavya Chopra, Rishabh Tiwari, Kurt Keutzer, Aditya Parameswaran, Dan Klein, Kannan Ramchandran, Matei Zaharia, Joseph~E. Gonzalez, and Ion Stoica.
\newblock Why {Do} {Multi}-{Agent} {LLM} {Systems} {Fail}?, March 2025.
\newblock URL \url{https://arxiv.org/abs/2503.13657v3}.

\bibitem[Chen et~al.(2023)Chen, Kansal, Chen, Jin, Reisler, Kim, and Rajpurkar]{chen_multimodal_2023}
Emma Chen, Aman Kansal, Julie Chen, Boyang~Tom Jin, Julia Reisler, David~E Kim, and Pranav Rajpurkar.
\newblock Multimodal {Clinical} {Benchmark} for {Emergency} {Care} ({MC}-{BEC}): {A} {Comprehensive} {Benchmark} for {Evaluating} {Foundation} {Models} in {Emergency} {Medicine}.
\newblock In A.~Oh, T.~Naumann, A.~Globerson, K.~Saenko, M.~Hardt, and S.~Levine, editors, \emph{Advances in {Neural} {Information} {Processing} {Systems}}, volume~36, pages 45794--45811. Curran Associates, Inc., 2023.
\newblock URL \url{https://proceedings.neurips.cc/paper_files/paper/2023/file/8f61049e8fe5b9ed714860b951066f1e-Paper-Datasets_and_Benchmarks.pdf}.

\bibitem[Chen et~al.(2024{\natexlab{a}})Chen, Wen, Pokojovy, Tseng, McCaffrey, Vo, Walser, and Moen]{chen_multi-modal_2024}
Junde Chen, Yuxin Wen, Michael Pokojovy, Tzu-Liang~(Bill) Tseng, Peter McCaffrey, Alexander Vo, Eric Walser, and Scott Moen.
\newblock Multi-modal learning for inpatient length of stay prediction.
\newblock \emph{Computers in Biology and Medicine}, 171:\penalty0 108121, March 2024{\natexlab{a}}.
\newblock ISSN 0010-4825.
\newblock \doi{10.1016/j.compbiomed.2024.108121}.
\newblock URL \url{https://www.sciencedirect.com/science/article/pii/S0010482524002051}.

\bibitem[Chen et~al.(2024{\natexlab{b}})Chen, Gui, Ouyang, Gao, Chen, Chen, Wang, Zhang, Cai, Ji, Yu, Wan, and Wang]{chen_huatuogpt-vision_2024}
Junying Chen, Chi Gui, Ruyi Ouyang, Anningzhe Gao, Shunian Chen, Guiming~Hardy Chen, Xidong Wang, Ruifei Zhang, Zhenyang Cai, Ke~Ji, Guangjun Yu, Xiang Wan, and Benyou Wang.
\newblock {HuatuoGPT}-{Vision}, {Towards} {Injecting} {Medical} {Visual} {Knowledge} into {Multimodal} {LLMs} at {Scale}, September 2024{\natexlab{b}}.
\newblock URL \url{http://arxiv.org/abs/2406.19280}.
\newblock arXiv:2406.19280 [cs].

\bibitem[Chen et~al.(2025)Chen, Wang, Cao, Liu, Gao, Cui, Zhu, Ye, Tian, Liu, Gu, Wang, Li, Ren, Chen, Luo, Wang, Jiang, Wang, He, Shi, Zhang, Lv, Wang, Shao, Chu, Tu, He, Wu, Deng, Ge, Chen, Zhang, Wang, Dou, Lu, Zhu, Lu, Lin, Qiao, Dai, and Wang]{chen_expanding_2025}
Zhe Chen, Weiyun Wang, Yue Cao, Yangzhou Liu, Zhangwei Gao, Erfei Cui, Jinguo Zhu, Shenglong Ye, Hao Tian, Zhaoyang Liu, Lixin Gu, Xuehui Wang, Qingyun Li, Yiming Ren, Zixuan Chen, Jiapeng Luo, Jiahao Wang, Tan Jiang, Bo~Wang, Conghui He, Botian Shi, Xingcheng Zhang, Han Lv, Yi~Wang, Wenqi Shao, Pei Chu, Zhongying Tu, Tong He, Zhiyong Wu, Huipeng Deng, Jiaye Ge, Kai Chen, Kaipeng Zhang, Limin Wang, Min Dou, Lewei Lu, Xizhou Zhu, Tong Lu, Dahua Lin, Yu~Qiao, Jifeng Dai, and Wenhai Wang.
\newblock Expanding {Performance} {Boundaries} of {Open}-{Source} {Multimodal} {Models} with {Model}, {Data}, and {Test}-{Time} {Scaling}, September 2025.
\newblock URL \url{http://arxiv.org/abs/2412.05271}.
\newblock arXiv:2412.05271 [cs].

\bibitem[Du et~al.(2024)Du, Li, Torralba, Tenenbaum, and Mordatch]{du_improving_2024}
Yilun Du, Shuang Li, Antonio Torralba, Joshua~B. Tenenbaum, and Igor Mordatch.
\newblock Improving factuality and reasoning in language models through multiagent debate.
\newblock In \emph{Proceedings of the 41st {International} {Conference} on {Machine} {Learning}}, volume 235 of \emph{{ICML}'24}, pages 11733--11763, Vienna, Austria, 2024. JMLR.org.

\bibitem[Elsharief et~al.(2025)Elsharief, Shurrab, Jorf, Lopez, Geras, and Shamout]{elsharief_medmod_2025}
Shaza Elsharief, Saeed Shurrab, Baraa~Al Jorf, Leopoldo Julian~Lechuga Lopez, Krzysztof~J. Geras, and Farah~E. Shamout.
\newblock {MedMod}: {Multimodal} {Benchmark} for {Medical} {Prediction} {Tasks} with {Electronic} {Health} {Records} and {Chest} {X}-{Ray} {Scans}.
\newblock In \emph{Proceedings of the sixth {Conference} on {Health}, {Inference}, and {Learning}}, pages 781--803. PMLR, July 2025.
\newblock URL \url{https://proceedings.mlr.press/v287/elsharief25a.html}.

\bibitem[Gao et~al.(2025)Gao, Li, Liu, Yu, Wang, Lin, and Lai]{gao_single-agent_2025}
Mingyan Gao, Yanzi Li, Banruo Liu, Yifan Yu, Phillip Wang, Ching-Yu Lin, and Fan Lai.
\newblock Single-agent or {Multi}-agent {Systems}? {Why} {Not} {Both}?, May 2025.
\newblock URL \url{https://arxiv.org/abs/2505.18286v1}.

\bibitem[Hayat et~al.(2022)Hayat, Geras, and Shamout]{hayat_medfuse_2022}
Nasir Hayat, Krzysztof~J. Geras, and Farah~E. Shamout.
\newblock {MedFuse}: {Multi}-modal fusion with clinical time-series data and chest {X}-ray images.
\newblock In \emph{Proceedings of the 7th {Machine} {Learning} for {Healthcare} {Conference}}, pages 479--503. PMLR, December 2022.
\newblock URL \url{https://proceedings.mlr.press/v182/hayat22a.html}.
\newblock ISSN: 2640-3498.

\bibitem[Hong et~al.(2023)Hong, Zhuge, Chen, Zheng, Cheng, Zhang, Wang, Wang, Yau, Lin, Zhou, Ran, Xiao, Wu, and Schmidhuber]{hong_metagpt_2023}
Sirui Hong, Mingchen Zhuge, Jiaqi Chen, Xiawu Zheng, Yuheng Cheng, Ceyao Zhang, Jinlin Wang, Zili Wang, Steven Ka~Shing Yau, Zijuan Lin, Liyang Zhou, Chenyu Ran, Lingfeng Xiao, Chenglin Wu, and Jürgen Schmidhuber.
\newblock {MetaGPT}: {Meta} {Programming} for {A} {Multi}-{Agent} {Collaborative} {Framework}, August 2023.
\newblock URL \url{https://arxiv.org/abs/2308.00352v7}.

\bibitem[Hou et~al.(2022)Hou, Dong, Wang, Li, and Che]{hou_metaprompting_2022}
Yutai Hou, Hongyuan Dong, Xinghao Wang, Bohan Li, and Wanxiang Che.
\newblock {MetaPrompting}: {Learning} to {Learn} {Better} {Prompts}.
\newblock In Nicoletta Calzolari, Chu-Ren Huang, Hansaem Kim, James Pustejovsky, Leo Wanner, Key-Sun Choi, Pum-Mo Ryu, Hsin-Hsi Chen, Lucia Donatelli, Heng Ji, Sadao Kurohashi, Patrizia Paggio, Nianwen Xue, Seokhwan Kim, Younggyun Hahm, Zhong He, Tony~Kyungil Lee, Enrico Santus, Francis Bond, and Seung-Hoon Na, editors, \emph{Proceedings of the 29th {International} {Conference} on {Computational} {Linguistics}}, pages 3251--3262, Gyeongju, Republic of Korea, October 2022. International Committee on Computational Linguistics.
\newblock URL \url{https://aclanthology.org/2022.coling-1.287/}.

\bibitem[Huang et~al.(2019)Huang, Altosaar, and Ranganath]{huang_clinicalbert_2019}
Kexin Huang, Jaan Altosaar, and Rajesh Ranganath.
\newblock {ClinicalBERT}: {Modeling} {Clinical} {Notes} and {Predicting} {Hospital} {Readmission}, April 2019.
\newblock URL \url{https://ui.adsabs.harvard.edu/abs/2019arXiv190405342H}.
\newblock ADS Bibcode: 2019arXiv190405342H.

\bibitem[Jin et~al.(2025)Jin, Wang, Yang, Zhu, Wright, Huang, Khandekar, Wan, Ai, Wilbur, He, Taylor, Chen, and Lu]{jin_agentmd_2025}
Qiao Jin, Zhizheng Wang, Yifan Yang, Qingqing Zhu, Donald Wright, Thomas Huang, Nikhil Khandekar, Nicholas Wan, Xuguang Ai, W.~John Wilbur, Zhe He, R.~Andrew Taylor, Qingyu Chen, and Zhiyong Lu.
\newblock {AgentMD}: {Empowering} language agents for risk prediction with large-scale clinical tool learning.
\newblock \emph{Nature Communications}, 16\penalty0 (1):\penalty0 9377, October 2025.
\newblock ISSN 2041-1723.
\newblock \doi{10.1038/s41467-025-64430-x}.
\newblock URL \url{https://www.nature.com/articles/s41467-025-64430-x}.

\bibitem[Johnson et~al.(2023{\natexlab{a}})Johnson, Pollard, Horng, Celi, and Mark]{johnson_mimic-iv-note_2023}
Alistair Johnson, Tom Pollard, Steven Horng, Leo~Anthony Celi, and Roger Mark.
\newblock {MIMIC}-{IV}-{Note}: {Deidentified} free-text clinical notes, 2023{\natexlab{a}}.
\newblock URL \url{https://physionet.org/content/mimic-iv-note/2.2/}.

\bibitem[Johnson et~al.(2019)Johnson, Pollard, Berkowitz, Greenbaum, Lungren, Deng, Mark, and Horng]{johnson_mimic-cxr_2019}
Alistair E.~W. Johnson, Tom~J. Pollard, Seth~J. Berkowitz, Nathaniel~R. Greenbaum, Matthew~P. Lungren, Chih-ying Deng, Roger~G. Mark, and Steven Horng.
\newblock {MIMIC}-{CXR}, a de-identified publicly available database of chest radiographs with free-text reports.
\newblock \emph{Scientific Data}, 6\penalty0 (1):\penalty0 317, December 2019.
\newblock ISSN 2052-4463.
\newblock \doi{10.1038/s41597-019-0322-0}.
\newblock URL \url{https://www.nature.com/articles/s41597-019-0322-0}.
\newblock Publisher: Nature Publishing Group.

\bibitem[Johnson et~al.(2023{\natexlab{b}})Johnson, Bulgarelli, Shen, Gayles, Shammout, Horng, Pollard, Hao, Moody, Gow, Lehman, Celi, and Mark]{johnson_mimic-iv_2023}
Alistair E.~W. Johnson, Lucas Bulgarelli, Lu~Shen, Alvin Gayles, Ayad Shammout, Steven Horng, Tom~J. Pollard, Sicheng Hao, Benjamin Moody, Brian Gow, Li-wei~H. Lehman, Leo~A. Celi, and Roger~G. Mark.
\newblock {MIMIC}-{IV}, a freely accessible electronic health record dataset.
\newblock \emph{Scientific Data}, 10\penalty0 (1):\penalty0 1, January 2023{\natexlab{b}}.
\newblock ISSN 2052-4463.
\newblock \doi{10.1038/s41597-022-01899-x}.
\newblock URL \url{https://www.nature.com/articles/s41597-022-01899-x}.

\bibitem[Jorf and Shamout(2025)]{al_jorf_medpatch_2025}
Baraa~Al Jorf and Farah~E. Shamout.
\newblock Medpatch: Confidence-guided multi-stage fusion for multimodal clinical data.
\newblock In Monica Agrawal, Kaivalya Deshpande, Matthew Engelhard, Shalmali Joshi, Shengpu Tang, and Iñigo Urteaga, editors, \emph{Proceedings of the 10th Machine Learning for Healthcare Conference}, volume 298 of \emph{Proceedings of Machine Learning Research}. PMLR, 15--16 Aug 2025.
\newblock URL \url{https://proceedings.mlr.press/v298/jorf25a.html}.

\bibitem[Kaesberg et~al.(2025)Kaesberg, Becker, Wahle, Ruas, and Gipp]{kaesberg_voting_2025}
Lars~Benedikt Kaesberg, Jonas Becker, Jan~Philip Wahle, Terry Ruas, and Bela Gipp.
\newblock Voting or {Consensus}? {Decision}-{Making} in {Multi}-{Agent} {Debate}.
\newblock In Wanxiang Che, Joyce Nabende, Ekaterina Shutova, and Mohammad~Taher Pilehvar, editors, \emph{Findings of the {Association} for {Computational} {Linguistics}: {ACL} 2025}, pages 11640--11671, Vienna, Austria, July 2025. Association for Computational Linguistics.
\newblock ISBN 979-8-89176-256-5.
\newblock \doi{10.18653/v1/2025.findings-acl.606}.
\newblock URL \url{https://aclanthology.org/2025.findings-acl.606/}.

\bibitem[Kalpelbe et~al.(2025)Kalpelbe, Adaambiik, and Peng]{kalpelbe_vision_2025}
Beria~Chingnabe Kalpelbe, Angel~Gabriel Adaambiik, and Wei Peng.
\newblock Vision {Language} {Models} in {Medicine}, February 2025.
\newblock URL \url{http://arxiv.org/abs/2503.01863}.
\newblock arXiv:2503.01863 [cs].

\bibitem[Kara and Gunel(2025)]{kara_clinical_2025}
Kaan Kara and Tuba Gunel.
\newblock Clinical {Risk} {Computation} by {Large} {Language} {Models} {Using} {Validated} {Risk} {Scores}.
\newblock \emph{Journal of Medical Systems}, 49\penalty0 (1):\penalty0 121, September 2025.
\newblock ISSN 1573-689X.
\newblock \doi{10.1007/s10916-025-02261-5}.
\newblock URL \url{https://doi.org/10.1007/s10916-025-02261-5}.

\bibitem[Khader et~al.(2023)Khader, Kather, Müller-Franzes, Wang, Han, Tayebi~Arasteh, Hamesch, Bressem, Haarburger, Stegmaier, Kuhl, Nebelung, and Truhn]{khader_medical_2023}
Firas Khader, Jakob~Nikolas Kather, Gustav Müller-Franzes, Tianci Wang, Tianyu Han, Soroosh Tayebi~Arasteh, Karim Hamesch, Keno Bressem, Christoph Haarburger, Johannes Stegmaier, Christiane Kuhl, Sven Nebelung, and Daniel Truhn.
\newblock Medical transformer for multimodal survival prediction in intensive care: integration of imaging and non-imaging data.
\newblock \emph{Scientific Reports}, 13\penalty0 (1):\penalty0 10666, July 2023.
\newblock ISSN 2045-2322.
\newblock \doi{10.1038/s41598-023-37835-1}.
\newblock URL \url{https://www.nature.com/articles/s41598-023-37835-1}.

\bibitem[Kim et~al.(2024)Kim, Park, Jeong, Chan, Xu, McDuff, Lee, Ghassemi, Breazeal, and Park]{kim_mdagents_2024}
Yubin Kim, Chanwoo Park, Hyewon Jeong, Yik~Siu Chan, Xuhai Xu, Daniel McDuff, Hyeonhoon Lee, Marzyeh Ghassemi, Cynthia Breazeal, and Hae~Won Park.
\newblock {MDAgents}: an adaptive collaboration of {LLMs} for medical decision-making.
\newblock In \emph{Proceedings of the 38th {International} {Conference} on {Neural} {Information} {Processing} {Systems}}, volume~37 of \emph{{NIPS} '24}, pages 79410--79452, Red Hook, NY, USA, December 2024. Curran Associates Inc.
\newblock ISBN 979-8-3313-1438-5.

\bibitem[Kim et~al.(2025)Kim, Gu, Park, Park, Schmidgall, Heydari, Yan, Zhang, Zhuang, Liu, Malhotra, Liang, Park, Yang, Xu, Du, Patel, Althoff, McDuff, and Liu]{kim_towards_2025}
Yubin Kim, Ken Gu, Chanwoo Park, Chunjong Park, Samuel Schmidgall, A.~Ali Heydari, Yao Yan, Zhihan Zhang, Yuchen Zhuang, Yun Liu, Mark Malhotra, Paul~Pu Liang, Hae~Won Park, Yuzhe Yang, Xuhai Xu, Yilun Du, Shwetak Patel, Tim Althoff, Daniel McDuff, and Xin Liu.
\newblock Towards a {Science} of {Scaling} {Agent} {Systems}, December 2025.
\newblock URL \url{https://arxiv.org/abs/2512.08296v3}.

\bibitem[Lee et~al.(2020)Lee, Yoon, Kim, Kim, Kim, So, and Kang]{lee_biobert_2020}
Jinhyuk Lee, Wonjin Yoon, Sungdong Kim, Donghyeon Kim, Sunkyu Kim, Chan~Ho So, and Jaewoo Kang.
\newblock {BioBERT}: a pre-trained biomedical language representation model for biomedical text mining.
\newblock \emph{Bioinformatics}, 36\penalty0 (4):\penalty0 1234--1240, February 2020.
\newblock ISSN 1367-4803.
\newblock \doi{10.1093/bioinformatics/btz682}.
\newblock URL \url{https://doi.org/10.1093/bioinformatics/btz682}.

\bibitem[Lee et~al.(2023)Lee, Lee, Hahn, Hyun, Choi, Ahn, and Lee]{lee_learning_2023}
Kwanhyung Lee, Soojeong Lee, Sangchul Hahn, Heejung Hyun, Edward Choi, Byungeun Ahn, and Joohyung Lee.
\newblock Learning {Missing} {Modal} {Electronic} {Health} {Records} with {Unified} {Multi}-modal {Data} {Embedding} and {Modality}-{Aware} {Attention}.
\newblock In \emph{Proceedings of the 8th {Machine} {Learning} for {Healthcare} {Conference}}, pages 423--442. PMLR, December 2023.
\newblock URL \url{https://proceedings.mlr.press/v219/lee23a.html}.

\bibitem[Lee et~al.(2025)Lee, Cho, Lee, Kim, Nam, Lee, Suh, and Ko]{lee_prompt_2025}
Sujung Lee, Won~Ik Cho, Youngrong Lee, Duck~Ju Kim, Kyeng~Hyun Nam, Sangmin Lee, Jungyo Suh, and Taehoon Ko.
\newblock A prompt framework for enhancing {LLM}-based explainability of medical machine learning models: an intensive care unit application.
\newblock \emph{BMC Medical Informatics and Decision Making}, 25\penalty0 (1):\penalty0 430, November 2025.
\newblock ISSN 1472-6947.
\newblock \doi{10.1186/s12911-025-03239-6}.
\newblock URL \url{https://doi.org/10.1186/s12911-025-03239-6}.

\bibitem[Lewis et~al.(2020)Lewis, Perez, Piktus, Petroni, Karpukhin, Goyal, Küttler, Lewis, Yih, Rocktäschel, Riedel, and Kiela]{lewis_retrieval-augmented_2020}
Patrick Lewis, Ethan Perez, Aleksandra Piktus, Fabio Petroni, Vladimir Karpukhin, Naman Goyal, Heinrich Küttler, Mike Lewis, Wen-tau Yih, Tim Rocktäschel, Sebastian Riedel, and Douwe Kiela.
\newblock Retrieval-augmented generation for knowledge-intensive {NLP} tasks.
\newblock In \emph{Proceedings of the 34th {International} {Conference} on {Neural} {Information} {Processing} {Systems}}, {NIPS} '20, pages 9459--9474, Red Hook, NY, USA, 2020. Curran Associates Inc.
\newblock ISBN 978-1-7138-2954-6.
\newblock URL \url{https://dl.acm.org/doi/10.5555/3495724.3496517}.

\bibitem[Li et~al.(2023)Li, Wong, Zhang, Usuyama, Liu, Yang, Naumann, Poon, and Gao]{li_llava-med_2023}
Chunyuan Li, Cliff Wong, Sheng Zhang, Naoto Usuyama, Haotian Liu, Jianwei Yang, Tristan Naumann, Hoifung Poon, and Jianfeng Gao.
\newblock {LLaVA}-{Med}: {Training} a {Large} {Language}-and-{Vision} {Assistant} for {Biomedicine} in {One} {Day}, June 2023.
\newblock URL \url{http://arxiv.org/abs/2306.00890}.
\newblock arXiv:2306.00890 [cs].

\bibitem[Liang et~al.(2024)Liang, He, Jiao, Wang, Wang, Wang, Yang, Shi, and Tu]{liang_encouraging_2024}
Tian Liang, Zhiwei He, Wenxiang Jiao, Xing Wang, Yan Wang, Rui Wang, Yujiu Yang, Shuming Shi, and Zhaopeng Tu.
\newblock Encouraging {Divergent} {Thinking} in {Large} {Language} {Models} through {Multi}-{Agent} {Debate}.
\newblock In Yaser Al-Onaizan, Mohit Bansal, and Yun-Nung Chen, editors, \emph{Proceedings of the 2024 {Conference} on {Empirical} {Methods} in {Natural} {Language} {Processing}}, pages 17889--17904, Miami, Florida, USA, November 2024. Association for Computational Linguistics.
\newblock \doi{10.18653/v1/2024.emnlp-main.992}.
\newblock URL \url{https://aclanthology.org/2024.emnlp-main.992/}.

\bibitem[Lu et~al.(2025)Lu, Li, Xia, Hu, Zhao, Ma, Wei, Li, Duan, Zhao, Han, Li, Chen, Tang, Hou, Du, Zhou, Zhang, Ding, Li, Li, Hu, Gu, Yang, Wang, Sun, Wang, Sun, Huang, He, Shi, Zhang, Zheng, Jiang, Gao, Wu, Chen, Chen, Chen, Xu, Luo, and Zhang]{lu2025ovis25technicalreport}
Shiyin Lu, Yang Li, Yu~Xia, Yuwei Hu, Shanshan Zhao, Yanqing Ma, Zhichao Wei, Yinglun Li, Lunhao Duan, Jianshan Zhao, Yuxuan Han, Haijun Li, Wanying Chen, Junke Tang, Chengkun Hou, Zhixing Du, Tianli Zhou, Wenjie Zhang, Huping Ding, Jiahe Li, Wen Li, Gui Hu, Yiliang Gu, Siran Yang, Jiamang Wang, Hailong Sun, Yibo Wang, Hui Sun, Jinlong Huang, Yuping He, Shengze Shi, Weihong Zhang, Guodong Zheng, Junpeng Jiang, Sensen Gao, Yi-Feng Wu, Sijia Chen, Yuhui Chen, Qing-Guo Chen, Zhao Xu, Weihua Luo, and Kaifu Zhang.
\newblock Ovis2.5 technical report.
\newblock \emph{arXiv:2508.11737}, 2025.

\bibitem[Lukac et~al.(2025)Lukac, Turner, Vangala, Chin, Khalili, Shih, Sarkisian, Cheng, and Mafi]{lukac_ambient_2025}
Paul~J. Lukac, William Turner, Sitaram Vangala, Aaron~T. Chin, Joshua Khalili, Ya-Chen~Tina Shih, Catherine Sarkisian, Eric~M. Cheng, and John~N. Mafi.
\newblock Ambient {AI} {Scribes} in {Clinical} {Practice}: {A} {Randomized} {Trial}.
\newblock \emph{NEJM AI}, 2\penalty0 (12):\penalty0 AIoa2501000, November 2025.
\newblock \doi{10.1056/AIoa2501000}.
\newblock URL \url{https://ai.nejm.org/doi/full/10.1056/AIoa2501000}.
\newblock Publisher: Massachusetts Medical Society.

\bibitem[Madaan et~al.(2023)Madaan, Tandon, Gupta, Hallinan, Gao, Wiegreffe, Alon, Dziri, Prabhumoye, Yang, Gupta, Majumder, Hermann, Welleck, Yazdanbakhsh, and Clark]{madaan_self-refine_2023}
Aman Madaan, Niket Tandon, Prakhar Gupta, Skyler Hallinan, Luyu Gao, Sarah Wiegreffe, Uri Alon, Nouha Dziri, Shrimai Prabhumoye, Yiming Yang, Shashank Gupta, Bodhisattwa~Prasad Majumder, Katherine Hermann, Sean Welleck, Amir Yazdanbakhsh, and Peter Clark.
\newblock Self-{Refine}: {Iterative} {Refinement} with {Self}-{Feedback}, May 2023.
\newblock URL \url{http://arxiv.org/abs/2303.17651}.
\newblock arXiv:2303.17651 [cs].

\bibitem[Nori et~al.(2023)Nori, Lee, Zhang, Carignan, Edgar, Fusi, King, Larson, Li, Liu, Luo, McKinney, Ness, Poon, Qin, Usuyama, White, and Horvitz]{nori_can_2023}
Harsha Nori, Yin~Tat Lee, Sheng Zhang, Dean Carignan, Richard Edgar, Nicolo Fusi, Nicholas King, Jonathan Larson, Yuanzhi Li, Weishung Liu, Renqian Luo, Scott~Mayer McKinney, Robert~Osazuwa Ness, Hoifung Poon, Tao Qin, Naoto Usuyama, Chris White, and Eric Horvitz.
\newblock Can {Generalist} {Foundation} {Models} {Outcompete} {Special}-{Purpose} {Tuning}? {Case} {Study} in {Medicine}, November 2023.
\newblock URL \url{http://arxiv.org/abs/2311.16452}.
\newblock arXiv:2311.16452 [cs].

\bibitem[Sellergren et~al.(2025)Sellergren, Kazemzadeh, Jaroensri, Kiraly, Traverse, Kohlberger, Xu, Jamil, Hughes, Lau, Chen, Mahvar, Yatziv, Chen, Sterling, Baby, Baby, Lai, Schmidgall, Yang, Chen, Bjornsson, Reddy, Brush, Philbrick, Asiedu, Mezerreg, Hu, Yang, Tiwari, Jansen, Singh, Liu, Azizi, Kamath, Ferret, Pathak, Vieillard, Merhej, Perrin, Matejovicova, Ramé, Riviere, Rouillard, Mesnard, Cideron, Grill, Ramos, Yvinec, Casbon, Buchatskaya, Alayrac, Lepikhin, Feinberg, Borgeaud, Andreev, Hardin, Dadashi, Hussenot, Joulin, Bachem, Matias, Chou, Hassidim, Goel, Farabet, Barral, Warkentin, Shlens, Fleet, Cotruta, Sanseviero, Martins, Kirk, Rao, Shetty, Steiner, Kirmizibayrak, Pilgrim, Golden, and Yang]{sellergren_medgemma_2025}
Andrew Sellergren, Sahar Kazemzadeh, Tiam Jaroensri, Atilla Kiraly, Madeleine Traverse, Timo Kohlberger, Shawn Xu, Fayaz Jamil, Cían Hughes, Charles Lau, Justin Chen, Fereshteh Mahvar, Liron Yatziv, Tiffany Chen, Bram Sterling, Stefanie~Anna Baby, Susanna~Maria Baby, Jeremy Lai, Samuel Schmidgall, Lu~Yang, Kejia Chen, Per Bjornsson, Shashir Reddy, Ryan Brush, Kenneth Philbrick, Mercy Asiedu, Ines Mezerreg, Howard Hu, Howard Yang, Richa Tiwari, Sunny Jansen, Preeti Singh, Yun Liu, Shekoofeh Azizi, Aishwarya Kamath, Johan Ferret, Shreya Pathak, Nino Vieillard, Ramona Merhej, Sarah Perrin, Tatiana Matejovicova, Alexandre Ramé, Morgane Riviere, Louis Rouillard, Thomas Mesnard, Geoffrey Cideron, Jean-bastien Grill, Sabela Ramos, Edouard Yvinec, Michelle Casbon, Elena Buchatskaya, Jean-Baptiste Alayrac, Dmitry Lepikhin, Vlad Feinberg, Sebastian Borgeaud, Alek Andreev, Cassidy Hardin, Robert Dadashi, Léonard Hussenot, Armand Joulin, Olivier Bachem, Yossi Matias, Katherine Chou, Avinatan Hassidim, Kavi Goel,
  Clement Farabet, Joelle Barral, Tris Warkentin, Jonathon Shlens, David Fleet, Victor Cotruta, Omar Sanseviero, Gus Martins, Phoebe Kirk, Anand Rao, Shravya Shetty, David~F. Steiner, Can Kirmizibayrak, Rory Pilgrim, Daniel Golden, and Lin Yang.
\newblock {MedGemma} {Technical} {Report}, July 2025.
\newblock URL \url{http://arxiv.org/abs/2507.05201}.
\newblock arXiv:2507.05201 [cs].

\bibitem[Shuaib(2024)]{shuaib_transforming_2024}
Ali Shuaib.
\newblock Transforming {Healthcare} with {AI}: {Promises}, {Pitfalls}, and {Pathways} {Forward}.
\newblock \emph{International Journal of General Medicine}, 17:\penalty0 1765--1771, May 2024.
\newblock \doi{10.2147/IJGM.S449598}.
\newblock URL \url{https://www.dovepress.com/transforming-healthcare-with-ai-promises-pitfalls-and-pathways-forward-peer-reviewed-fulltext-article-IJGM}.
\newblock Publisher: Dove Press.

\bibitem[Singhal et~al.(2023)Singhal, Azizi, Tu, Mahdavi, Wei, Chung, Scales, Tanwani, Cole-Lewis, Pfohl, Payne, Seneviratne, Gamble, Kelly, Babiker, Schärli, Chowdhery, Mansfield, Demner-Fushman, Agüera~y Arcas, Webster, Corrado, Matias, Chou, Gottweis, Tomasev, Liu, Rajkomar, Barral, Semturs, Karthikesalingam, and Natarajan]{singhal_large_2023}
Karan Singhal, Shekoofeh Azizi, Tao Tu, S.~Sara Mahdavi, Jason Wei, Hyung~Won Chung, Nathan Scales, Ajay Tanwani, Heather Cole-Lewis, Stephen Pfohl, Perry Payne, Martin Seneviratne, Paul Gamble, Chris Kelly, Abubakr Babiker, Nathanael Schärli, Aakanksha Chowdhery, Philip Mansfield, Dina Demner-Fushman, Blaise Agüera~y Arcas, Dale Webster, Greg~S. Corrado, Yossi Matias, Katherine Chou, Juraj Gottweis, Nenad Tomasev, Yun Liu, Alvin Rajkomar, Joelle Barral, Christopher Semturs, Alan Karthikesalingam, and Vivek Natarajan.
\newblock Large language models encode clinical knowledge.
\newblock \emph{Nature}, 620\penalty0 (7972):\penalty0 172--180, August 2023.
\newblock ISSN 1476-4687.
\newblock \doi{10.1038/s41586-023-06291-2}.
\newblock URL \url{https://www.nature.com/articles/s41586-023-06291-2}.

\bibitem[Singhal et~al.(2025)Singhal, Tu, Gottweis, Sayres, Wulczyn, Amin, Hou, Clark, Pfohl, Cole-Lewis, Neal, Rashid, Schaekermann, Wang, Dash, Chen, Shah, Lachgar, Mansfield, Prakash, Green, Dominowska, Agüera~y Arcas, Tomašev, Liu, Wong, Semturs, Mahdavi, Barral, Webster, Corrado, Matias, Azizi, Karthikesalingam, and Natarajan]{singhal_toward_2025}
Karan Singhal, Tao Tu, Juraj Gottweis, Rory Sayres, Ellery Wulczyn, Mohamed Amin, Le~Hou, Kevin Clark, Stephen~R. Pfohl, Heather Cole-Lewis, Darlene Neal, Qazi~Mamunur Rashid, Mike Schaekermann, Amy Wang, Dev Dash, Jonathan~H. Chen, Nigam~H. Shah, Sami Lachgar, Philip~Andrew Mansfield, Sushant Prakash, Bradley Green, Ewa Dominowska, Blaise Agüera~y Arcas, Nenad Tomašev, Yun Liu, Renee Wong, Christopher Semturs, S.~Sara Mahdavi, Joelle~K. Barral, Dale~R. Webster, Greg~S. Corrado, Yossi Matias, Shekoofeh Azizi, Alan Karthikesalingam, and Vivek Natarajan.
\newblock Toward expert-level medical question answering with large language models.
\newblock \emph{Nature Medicine}, 31\penalty0 (3):\penalty0 943--950, March 2025.
\newblock ISSN 1546-170X.
\newblock \doi{10.1038/s41591-024-03423-7}.
\newblock URL \url{https://www.nature.com/articles/s41591-024-03423-7}.
\newblock Publisher: Nature Publishing Group.

\bibitem[Tai-Seale et~al.(2024)Tai-Seale, Baxter, Vaida, Walker, Sitapati, Osborne, Diaz, Desai, Webb, Polston, Helsten, Gross, Thackaberry, Mandvi, Lillie, Li, Gin, Achar, Hofflich, Sharp, Millen, and Longhurst]{tai-seale_ai-generated_2024}
Ming Tai-Seale, Sally~L. Baxter, Florin Vaida, Amanda Walker, Amy~M. Sitapati, Chad Osborne, Joseph Diaz, Nimit Desai, Sophie Webb, Gregory Polston, Teresa Helsten, Erin Gross, Jessica Thackaberry, Ammar Mandvi, Dustin Lillie, Steve Li, Geneen Gin, Suraj Achar, Heather Hofflich, Christopher Sharp, Marlene Millen, and Christopher~A. Longhurst.
\newblock {AI}-{Generated} {Draft} {Replies} {Integrated} {Into} {Health} {Records} and {Physicians}’ {Electronic} {Communication}.
\newblock \emph{JAMA Network Open}, 7\penalty0 (4):\penalty0 e246565, April 2024.
\newblock ISSN 2574-3805.
\newblock \doi{10.1001/jamanetworkopen.2024.6565}.
\newblock URL \url{https://doi.org/10.1001/jamanetworkopen.2024.6565}.

\bibitem[Tan et~al.(2024)Tan, Merrill, Gupta, Althoff, and Hartvigsen]{tan_are_2024}
Mingtian Tan, Mike~A. Merrill, Vinayak Gupta, Tim Althoff, and Thomas Hartvigsen.
\newblock Are language models actually useful for time series forecasting?
\newblock In \emph{Proceedings of the 38th {International} {Conference} on {Neural} {Information} {Processing} {Systems}}, volume~37 of \emph{{NIPS} '24}, pages 60162--60191, Red Hook, NY, USA, December 2024. Curran Associates Inc.
\newblock ISBN 979-8-3313-1438-5.

\bibitem[Tang et~al.(2024)Tang, Zou, Zhang, Li, Zhao, Zhang, Cohan, and Gerstein]{tang_medagents_2024}
Xiangru Tang, Anni Zou, Zhuosheng Zhang, Ziming Li, Yilun Zhao, Xingyao Zhang, Arman Cohan, and Mark Gerstein.
\newblock {MedAgents}: {Large} {Language} {Models} as {Collaborators} for {Zero}-shot {Medical} {Reasoning}.
\newblock In Lun-Wei Ku, Andre Martins, and Vivek Srikumar, editors, \emph{Findings of the {Association} for {Computational} {Linguistics}: {ACL} 2024}, pages 599--621, Bangkok, Thailand, August 2024. Association for Computational Linguistics.
\newblock \doi{10.18653/v1/2024.findings-acl.33}.
\newblock URL \url{https://aclanthology.org/2024.findings-acl.33/}.

\bibitem[von Eschenbach(2021)]{von_eschenbach_transparency_2021}
Warren~J. von Eschenbach.
\newblock Transparency and the {Black} {Box} {Problem}: {Why} {We} {Do} {Not} {Trust} {AI}.
\newblock \emph{Philosophy \& Technology}, 34\penalty0 (4):\penalty0 1607--1622, December 2021.
\newblock ISSN 2210-5441.
\newblock \doi{10.1007/s13347-021-00477-0}.
\newblock URL \url{https://doi.org/10.1007/s13347-021-00477-0}.

\bibitem[Wang et~al.(2022)Wang, Wei, Schuurmans, Le, Chi, Narang, Chowdhery, and Zhou]{wang_self-consistency_2022}
Xuezhi Wang, Jason Wei, Dale Schuurmans, Quoc Le, Ed~Chi, Sharan Narang, Aakanksha Chowdhery, and Denny Zhou.
\newblock Self-{Consistency} {Improves} {Chain} of {Thought} {Reasoning} in {Language} {Models}, March 2022.
\newblock URL \url{https://arxiv.org/abs/2203.11171v4}.

\bibitem[Wei et~al.(2022)Wei, Wang, Schuurmans, Bosma, Ichter, Xia, Chi, Le, and Zhou]{wei_chain--thought_2022}
Jason Wei, Xuezhi Wang, Dale Schuurmans, Maarten Bosma, Brian Ichter, Fei Xia, Ed~H. Chi, Quoc~V. Le, and Denny Zhou.
\newblock Chain-of-thought prompting elicits reasoning in large language models.
\newblock In \emph{Proceedings of the 36th {International} {Conference} on {Neural} {Information} {Processing} {Systems}}, {NIPS} '22, pages 24824--24837, Red Hook, NY, USA, November 2022. Curran Associates Inc.
\newblock ISBN 978-1-7138-7108-8.

\bibitem[Yang et~al.(2025)Yang, Chai, Shao, Song, Qi, Rui, and Zhang]{yang_agentnet_2025}
Yingxuan Yang, Huacan Chai, Shuai Shao, Yuanyi Song, Siyuan Qi, Renting Rui, and Weinan Zhang.
\newblock {AgentNet}: {Decentralized} {Evolutionary} {Coordination} for {LLM}-based {Multi}-{Agent} {Systems}, April 2025.
\newblock URL \url{https://arxiv.org/abs/2504.00587v2}.

\bibitem[Zeng et~al.(2025)Zeng, Fu, Zhou, Yu, Liu, Wen, Thompson, Etzioni, and Yetisgen]{zeng_traj-coa_nodate}
Sihang Zeng, Yujuan Fu, Sitong Zhou, Zixuan Yu, Lucas~Jing Liu, Jun Wen, Matthew Thompson, Ruth Etzioni, and Meliha Yetisgen.
\newblock Traj-coa: Patient trajectory modeling via chain-of-agents for lung cancer risk prediction, 2025.
\newblock URL \url{https://arxiv.org/abs/2510.10454}.
\newblock Accepted by NeurIPS 2025 GenAI4Health Workshop.

\bibitem[Zhu et~al.(2025)Zhu, He, Hu, Zheng, Zhang, Wang, Gao, Ma, and Yu]{zhu_medagentboard_2025}
Yinghao Zhu, Ziyi He, Haoran Hu, Xiaochen Zheng, Xichen Zhang, Zixiang Wang, Junyi Gao, Liantao Ma, and Lequan Yu.
\newblock {MedAgentBoard}: {Benchmarking} {Multi}-{Agent} {Collaboration} with {Conventional} {Methods} for {Diverse} {Medical} {Tasks}.
\newblock October 2025.
\newblock URL \url{https://openreview.net/forum?id=BPpG4qQaNj}.

\end{thebibliography}

\onecolumn

% \appendix

% \numberwithin{figure}{section}
% \numberwithin{table}{section}
% \numberwithin{algocf}{section}

% \renewcommand{\thefigure}{\thesection\arabic{figure}}
% \renewcommand{\thetable}{\thesection\arabic{table}}
% \renewcommand{\thealgocf}{\thesection\arabic{algocf}}

% \renewcommand{\thefigure}{\thesection\arabic{figure}}
% \renewcommand{\thetable}{\thesection\arabic{table}}
% \renewcommand{\thealgocf}{\thesection\arabic{algocf}}

\appendix

% Make appendix figures and tables share one counter, reset by appendix section.
\numberwithin{figure}{section}

\makeatletter
\let\c@table\c@figure
\makeatother

\renewcommand{\thefigure}{\thesection\arabic{figure}}
\renewcommand{\thetable}{\thesection\arabic{figure}}

% Keep algorithms separate, if desired.
\numberwithin{algocf}{section}
\renewcommand{\thealgocf}{\thesection\arabic{algocf}}

\section{Algorithms}
This section shows all the algorithms used for each agentic setup. Single agents follow algorithm \ref{alg:single_unimodal} in the unimodal setup and algorithm \ref{alg:single_multimodal} in the multimodal setup. Multi-agent frameworks follow algorithm \ref{alg:multi_agent}.
\begin{algorithm}[h]
    \hrule
    \vspace{0.5em}
    \caption{Single Agent Unimodal (PS Only)}
    \label{alg:single_unimodal}
    \vspace{0.5em}
    \hrule
    \vspace{0.4em}
    \DontPrintSemicolon
    \KwIn{Patient Summary $\mathbf{x}_{ps}$,
    System Prompt $\mathcal{P}_{task}$}
    \KwOut{Prediction $\hat{y}$}
    \BlankLine
    $Context \gets \mathrm{concat}(\mathcal{P}_{task}, \mathbf{x}_{ps})$\; 
    \BlankLine
    $\hat{y} \gets \mathcal{A}(Context)$\;
    \BlankLine
    \Return{$\hat{y}$}\;
    \vspace{0.5em}
    \hrule
\end{algorithm}

\begin{algorithm}[h!]
    \hrule
    \vspace{0.5em}
    \caption{Single Agent Multimodal}
    \label{alg:single_multimodal}
    \vspace{0.5em}
    \hrule
    \vspace{0.4em}
    \DontPrintSemicolon
    \KwIn{Available modalities $\mathcal{X}_{available}$, \\
    System Prompt $\mathcal{P}_{task}$}
    \KwOut{Prediction $\hat{y}$}
    \BlankLine
    $Context \gets \mathcal{P}_{task}$\;
    \BlankLine
    \ForEach{$\mathbf{x}_m \in \mathcal{X}_{available}$}{
        $Context \gets \mathrm{concat}(Context, \text{``Modality } m \text{: "}, \mathbf{x}_m)$\;
    }
    \BlankLine
    $\hat{y} \gets \mathcal{A}(Context)$\;
    \BlankLine
    \Return{$\hat{y}$}\;
    \vspace{0.5em}
    \hrule
\end{algorithm}

\begin{algorithm}[h]
    \hrule
    \vspace{0.5em}
    \caption{Multi-Agent Multimodal}
    \label{alg:multi_agent}
    \vspace{0.5em}
    \hrule
    \vspace{0.4em}
    \DontPrintSemicolon
    \KwIn{Available modalities $\mathcal{X}_{available}$, \\
    System Prompt $\mathcal{P}_{task}$}
    \KwOut{Prediction $\hat{y}$}
    \BlankLine
    $S \gets 0$\;
    \BlankLine
    \ForEach{$\mathbf{x}_m \in \mathcal{X}_{available}$}{
        $Context_m \gets \mathrm{concat}(\mathcal{P}_{task}, \mathbf{x}_m)$\;
        \BlankLine
        $p_m \gets \mathcal{A}_m(Context_m)$\;
        \BlankLine
        $S \gets S + p_m$\;
    }
    \BlankLine
    $\bar{p} \gets S / |\mathcal{X}_{available}|$\;
    \BlankLine
    $\hat{y} \gets \mathbb{I}(\bar{p} > 0.5)$\;
    \BlankLine
    \Return{$\hat{y}$}\;
    \vspace{0.5em}
    \hrule
\end{algorithm}
\newpage

\section{Additional Results}\label{apd:second}

\subsection{Intern Multimodal}
\begin{figure*}[h]
    \centering
    \includegraphics[width=0.75\textwidth]{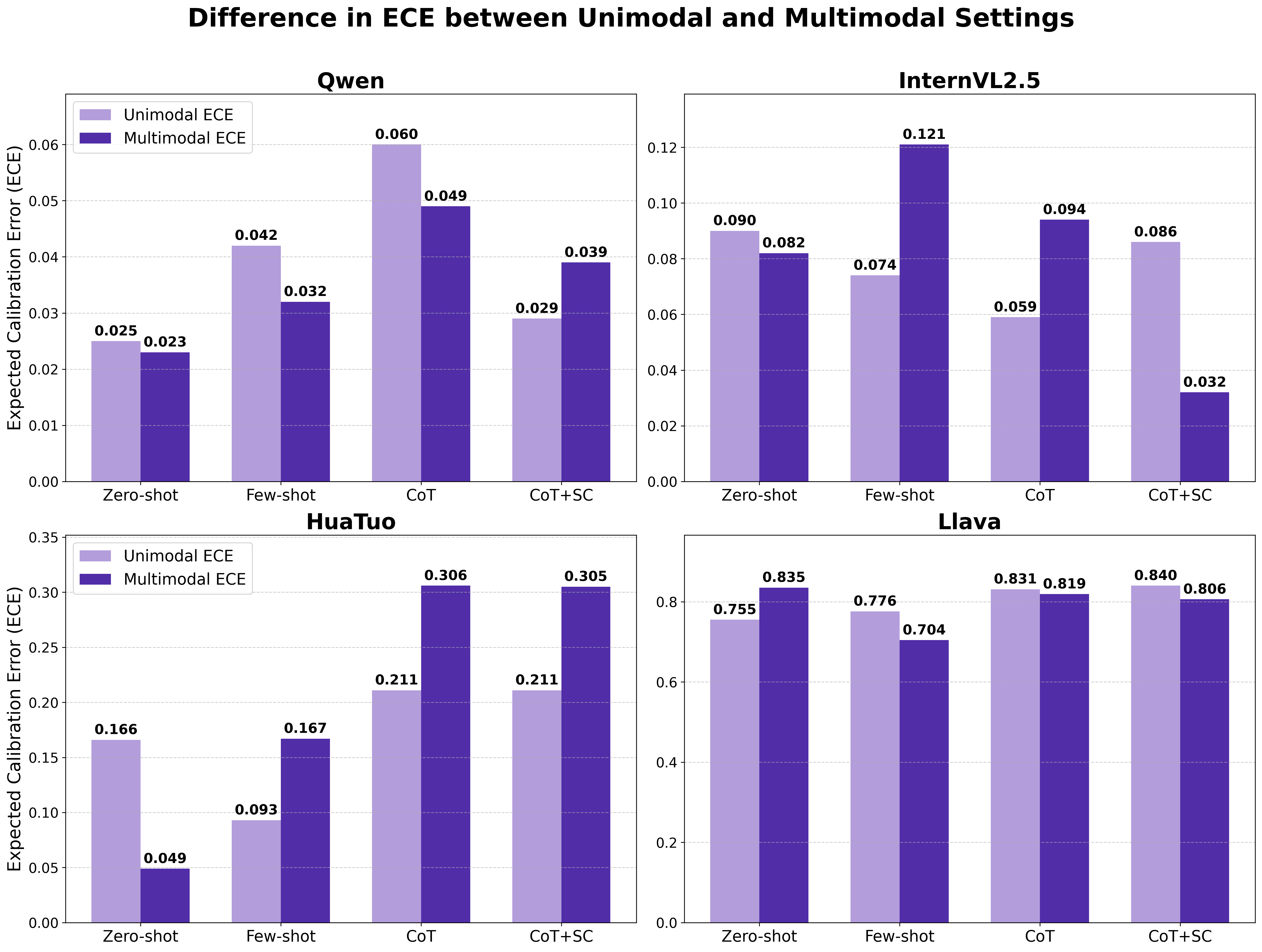} \vspace{-5mm}
    \caption{Bar graphs showing the difference in ECE between the unimodal and multimodal settings for the single agent frameworks across the four backbones.}\vspace{-5mm}
    \label{calibration}
\end{figure*}

Table \ref{tab:intern_multimodal} details the performance of the InternVL2.5 model. Note that this architecture supports three modalities (PS, CXR, RR), excluding EHR. We report the Area Under the Receiver Operating Characteristic (AUROC), Area Under the Precision-Recall Curve (AUPRC), and Expected Calibration Error (ECE) for both In-Hospital Mortality and Length of Stay ($>7$ Days) prediction tasks. Figure \ref{calibration} illustrates the shift in ECE when transitioning from the unimodal to the multimodal setting within single-agent frameworks. Consistent with our findings, the chart highlights that adding modalities generally improves calibration for robust backbones, though this benefit varies across model architectures.

\begin{table}[h]
\caption{\textbf{InternVL2.5 Multimodal Results.} Results for InternVL2.5 are reported separately as this model supports only 3 modalities (missing Timeseries). Best overall performance in each column is bolded.}
\label{tab:intern_multimodal}
\centering
\resizebox{\linewidth}{!}{%
\begin{tabular}{lllcccccc}
\toprule
\multirow{2}{*}{\textbf{Backbone}} & \multirow{2}{*}{\textbf{Arch.}} & \multirow{2}{*}{\textbf{Method}} & \multicolumn{3}{c}{\textbf{In-Hospital Mortality}} & \multicolumn{3}{c}{\textbf{Length of Stay ($>$7 Days)}} \\
\cmidrule(lr){4-6} \cmidrule(lr){7-9}
 & & & \textbf{AUROC} & \textbf{AUPRC} & \textbf{ECE} & \textbf{AUROC} & \textbf{AUPRC} & \textbf{ECE} \\
\midrule
\multirow{8}{*}{\textbf{Intern}} & \multirow{4}{*}{Single} 
 & Zero-shot & 0.700 (0.680 - 0.723) & 0.247 (0.223 - 0.281) & 0.082 & \textbf{0.721 (0.703 - 0.738)} & \textbf{0.358 (0.331 - 0.387)} & 0.354 \\
 & & Few-shot & \textbf{0.745 (0.724 - 0.766)} & \textbf{0.282 (0.252 - 0.315)} & 0.121 & 0.677 (0.658 - 0.695) & 0.302 (0.280 - 0.328) & 0.232 \\
 & & CoT  & 0.683 (0.662 - 0.705) & 0.216 (0.194 - 0.242) & 0.094  & 0.674 (0.655 - 0.693) & 0.319 (0.293 - 0.348) & 0.445 \\
 & & CoT-SC & 0.727 (0.707-0.748) & 0.270 (0.707 - 0.748) & \textbf{0.032} & 0.690 (0.670 - 0.709) & 0.345 (0.319 - 0.377) & 0.456 \\
 \cmidrule{2-9}
 & \multirow{4}{*}{Multi} 
 & Majority Vote & 0.709 (0.688 - 0.731) & 0.249 (0.220 - 0.282) & 0.106 & 0.679 (0.658 - 0.697) & 0.306 (0.281 - 0.335) & 0.056 \\
 & & Debate (Uni) & 0.627 (0.603 - 0.651) & 0.181 (0.161 0.207) & 0.053 & 0.641 (0.620 - 0.660) & 0.285 (0.259 - 0.313) & \textbf{0.006} \\
 & & Meta-Prompt & 0.554 (0.529 - 0.580) & 0.142 (0.128 - 0.159) & 0.223 & 0.550 (0.528 - 0.571) & 0.230 (0.212 - 0.251) & 0.374 \\
\bottomrule
\end{tabular}}
\end{table}

\newpage

\section{Additional Ablations}\label{apd:third}

\subsection{Modality Addition Ablations}
\begin{figure*}[h]
    \centering
    \includegraphics[width=0.8\textwidth]{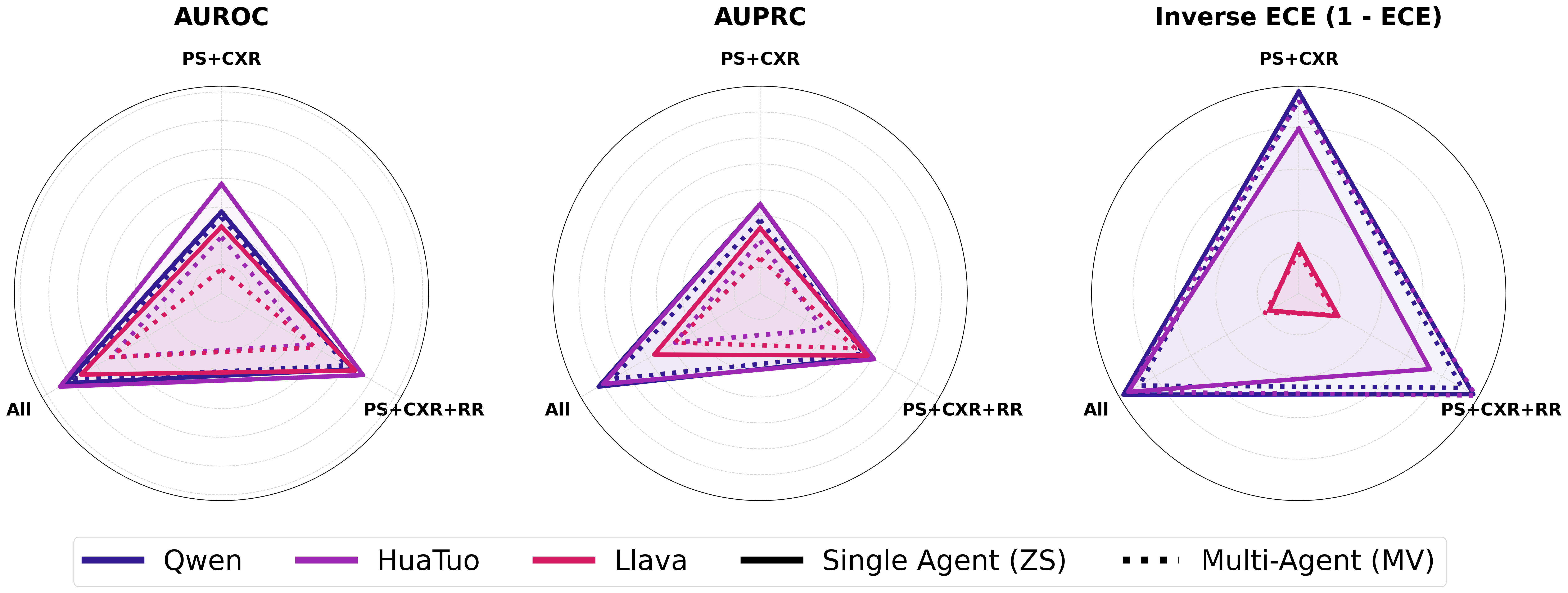} \vspace{-3mm}
    \caption{Radar charts showing how the single agent setup benefits more from the multimodality than the multiagent setup. Each circle corresponds to a metric. The dotted lines correspond to majority vote and the solid ones to zero-shot. Note the the chart shows the inverse of ECE for consistency with the other metrics.}\vspace{-5mm}
    \label{Figure3}
\end{figure*}

In this section, we provide visual support for the calibration and performance trends observed in our ablation studies. Figure \ref{Figure3} offers a holistic comparison between the effect of adding more modalities to the single agent and multi-agent setups. The improvement of the single agent configuration as more modalities are added compared to the drop in majority vote reinforces our conclusion that unified context processing synthesizes heterogeneous clinical data more effectively than decentralized mechanisms.

\subsection{Weighted Majority Vote}

We examine the effect of weighting each modality agent by its unimodal AUROC performance in the Majority vote backbone. The results are shown in table \ref{tab:WMV_performance}
\begin{table*}[ht]
\caption{\textbf{Performance Results.} Comparison of Unimodal baselines and Multimodal Multi-Agent voting methods. Best overall performance in each column is bolded.}
\label{tab:WMV_performance}
\centering
\resizebox{\linewidth}{!}{%
\begin{tabular}{lllcc}
\toprule
\multirow{2}{*}{\textbf{Backbone}} & \multirow{2}{*}{\textbf{Arch.}} & \multirow{2}{*}{\textbf{Method}} & \multicolumn{2}{c}{\textbf{Performance}} \\
\cmidrule(lr){4-5}
 & & & \textbf{AUROC} & \textbf{AUPRC} \\
\midrule
\multirow{6}{*}{\textbf{Qwen}} 
 & \multirow{4}{*}{Unimodal} 
 & PS & 0.666 & 0.232 \\
 & & RR & 0.684 & 0.222 \\
 & & EHR & 0.662 & 0.230 \\
 & & CXR & 0.538 & 0.141 \\
 \cmidrule{2-5}
 & \multirow{2}{*}{Multimodal Multi-Agent} 
 & Majority Vote & 0.748 & 0.315 \\
 & & Weighted Majority Vote & \textbf{0.750} & \textbf{0.318} \\
 \bottomrule 
\end{tabular}%
}
\end{table*}
\newpage
\subsection{Sycophancy Analysis}
We examine the reasoning traces from the debate between unimodal agents baseline and summarize our findings in Table \ref{tab:sycophancy}. Notably, the agents seem to exhibit varying levels of sycophancy and echo chamber agreement depending on the backbone used.
\begin{table*}[h]
\centering
\caption{Debate traces analysis and echo chamber behavior per backbone.}
\label{tab:sycophancy}
\begin{tabular}{llp{8cm}}
\toprule
Backbone & Pattern & Evidence \\
\midrule
Qwen     & Premature agreement & Consensus in round 1 for all 4,925 cases. \\
Huatuo   & Weak echo chamber & Some later consensuses aligned with the initial majority. \\
MedGemma & Strong echo chamber & 75/76 initial disagreements later converged. All followed the round-1 majority. \\
LLaVaMed & Unstable communication & Disagreements rarely resolved into stable consensus. \\
\bottomrule
\end{tabular}
\end{table*}

\subsection{Debate Between Multimodal Agents}
In Table \ref{tab:debate_multimodal}, we examine the performance of debate between multimodal agents. Each agent in this setup receives the full multimodal context and debates with the other multimodal agents until consensus.

\begin{table*}[ht]
\caption{\textbf{Debate Multimodal Results.} Comparison of in-hospital mortality and length of stay prediction performance for HuaTuo and Qwen using the debate architecture with multimodal agents. Best overall performance in each column is bolded.}
\label{tab:debate_multimodal}
\centering
\resizebox{\linewidth}{!}{%
\begin{tabular}{lcccc}
\toprule
\multirow{2}{*}{\textbf{Backbone}} & \multicolumn{2}{c}{\textbf{In-Hospital Mortality}} & \multicolumn{2}{c}{\textbf{Length of Stay ($>$7 Days)}} \\
\cmidrule(lr){2-3} \cmidrule(lr){4-5}
 & \textbf{AUROC} & \textbf{AUPRC} & \textbf{AUROC} & \textbf{AUPRC} \\
\midrule
\textbf{HuaTuo} & 0.684 (0.663 - 0.707) & 0.235 (0.210 - 0.266) & \textbf{0.592 (0.572 - 0.613)} & \textbf{0.249 (0.231 - 0.272)} \\
\textbf{Qwen} & \textbf{0.707 (0.686 - 0.728)} & \textbf{0.243 (0.215 - 0.275)} & 0.561 (0.541 - 0.581) & 0.231 (0.213 - 0.252) \\
\bottomrule
\end{tabular}%
}
\end{table*}

\subsection{Note-based Task}
In Table \ref{tab:qwen_ckd}, we test the performance of simple single agent and multi-agent setups on a note-based chronic kidney disease detection task. We extract the labels from MIMIC IV.
\begin{table*}[h]
\caption{\textbf{Multimodal Results.} Comparison of Unimodal, Multimodal, and Multi-Agent architectures for Chronic Kidney Disease Detection. Best overall performance in each column is bolded.}
\label{tab:qwen_ckd}
\centering
\resizebox{\linewidth}{!}{%
\begin{tabular}{lllcc}
\toprule
\multirow{2}{*}{\textbf{Backbone}} & \multirow{2}{*}{\textbf{Arch.}} & \multirow{2}{*}{\textbf{Method}} & \multicolumn{2}{c}{\textbf{Chronic Kidney Disease Detection}} \\
\cmidrule(lr){4-5}
 & & & \textbf{AUROC} & \textbf{AUPRC} \\
\midrule
\multirow{2}{*}{\textbf{Supervised}} 
 & Unimodal & BioBERT & 0.878 & 0.705 \\
 \cmidrule{2-5}
 & Multimodal & MedPatch & 0.807 & 0.532 \\
\midrule
\multirow{3}{*}{\textbf{Qwen}} 
 & Unimodal Single Agent & Zero-shot & \textbf{0.968} & \textbf{0.924} \\
 \cmidrule{2-5}
 & Multimodal Multi-Agent & Majority Vote & 0.959 & 0.897 \\
 \bottomrule 
\end{tabular}%
}
\end{table*}

\subsection{MedGemma}
Table \ref{tab:medgemma} shows the results on the in-hospital mortality task using the MedGemma 4B backbone \citep{sellergren_medgemma_2025}. We note that performance does not diverge much from the larger backbones.

\begin{table*}[ht]
\caption{\textbf{Multimodal Results.} Comparison of Unimodal Single-Agent, Multimodal Single-Agent, and Multimodal Multi-Agent frameworks on the mortality task using the MedGemma backbone. Best overall performance in each column is bolded.}
\label{tab:medgemma}
\centering
\resizebox{\linewidth}{!}{%
\begin{tabular}{lllcc}
\toprule
\multirow{2}{*}{\textbf{Backbone}} & \multirow{2}{*}{\textbf{Arch.}} & \multirow{2}{*}{\textbf{Method}} & \multicolumn{2}{c}{\textbf{In-Hospital Mortality}} \\
\cmidrule(lr){4-5}
 & & & \textbf{AUROC} & \textbf{AUPRC} \\
\midrule
\multirow{8}{*}{\textbf{MedGemma}} & \multirow{4}{*}{Unimodal Single Agent} 
 & Zero-shot & 0.626 (0.602 - 0.649) & 0.179 (0.160 - 0.201) \\
 & & Few-shot & 0.691 (0.207 - 0.714) & 0.232 (0.207 - 0.263) \\
 & & CoT & 0.624 (0.600 - 0.647) & 0.172 (0.154 - 0.190) \\
 & & Self-Refine & 0.628 (0.603 - 0.652) & 0.183 (0.163 - 0.206) \\
 \cmidrule{2-5}
 & \multirow{2}{*}{Multimodal Single Agent} 
 & Zero-shot & 0.735 (0.714 - 0.755) & 0.250 (0.225 - 0.282) \\
 & & Few-shot & \textbf{0.744 (0.723 - 0.765)} & \textbf{0.280 (0.250 - 0.314)} \\
 \cmidrule{2-5}
 & \multirow{2}{*}{Multimodal Multi-Agent} 
 & Majority Vote & 0.687 (0.664 - 0.708) & 0.226 (0.201 - 0.256) \\
 & & Debate & 0.594 (0.569 - 0.619) & 0.172 (0.151 - 0.197) \\
 \bottomrule 
\end{tabular}%
}
\end{table*}

\subsection{EHR Serialization}
Table \ref{tab:log_comparison} details an ablation on EHR serialization methods. We tested our standard log baseline against two reduced-context approaches: the summary method (aggregating extreme and average baseline values) and the delta method (representing only the net clinical trajectory). Best overall performance in each column is bolded.
\begin{table}[ht]
\caption{\textbf{Performance Results.} Comparison of Log, Summary, and Delta methods.}
\label{tab:log_comparison}
\centering
\begin{tabular}{lcc}
\toprule
\textbf{Method} & \textbf{AUROC} & \textbf{AUPRC} \\
\midrule
Original Log & 0.756 & \textbf{0.330} \\
Summary & 0.\textbf{760} & 0.301 \\
Delta & \textbf{0.760} & 0.303 \\
\bottomrule
\end{tabular}
\end{table}

\end{document}